\documentclass[twocolumn]{svjour3}          
\smartqed  
\usepackage{graphicx}
\usepackage{overpic}
\usepackage{caption}
\usepackage{amssymb}
\usepackage{amsmath}
\usepackage{bm}
\usepackage{multirow}
\newcommand{\comment}[1]{}
\DeclareMathOperator*{\argmin}{argmin}
\newcommand{\mat}[1]{\mathbf{#1}}
\renewcommand{\vec}[1]{\mathbf{#1}}
\providecommand{\mC}{\ensuremath{\mat{C}}}
\providecommand{\mD}{\ensuremath{\mat{D}}}
\providecommand{\mG}{\ensuremath{\mat{G}}}
\providecommand{\vc}{\ensuremath{\vec{c}}}
\providecommand{\vh}{\ensuremath{\vec{h}}}
\providecommand{\vx}{\ensuremath{\vec{x}}}
\providecommand{\vz}{\ensuremath{\vec{z}}}
\newcommand{\RR}{\mathbb{R}}
\begin{document}

\title{Two Birds with One Stone: Transforming and Generating Facial Images with Iterative GAN 
}


\author{Dan Ma      \and
        Bin Liu         \and
        Zhao Kang    \and
        Jiayu Zhou   \and
        Jianke Zhu  \and
        Zenglin Xu 
}


\institute{Dan Ma, Bin Liu, Zhao Kang, Zenglin Xu \at
              SMILE Lab, School of Computer Science and Engineering, University of Electronic Science and Technology of China, Chengdu, Sichuan, China \\
              \email{zlxu@uestc.edu.cn}           
        \and
           Jiayu Zhou \at
           Department of Computer Science and Engineering, Michigan State University, East Lansing, MI, United States \\
           \and
           Jianke Zhu \at
           College of Computer Science and Technology, Zhejiang University, Hangzhou, Zhejiang, China 
}

\date{Received: date / Accepted: date}

\maketitle

\begin{abstract}
Generating high fidelity identity-preserving faces with different facial attributes has a wide range of applications. Although a number of generative models have been developed to tackle this problem, there is still much room for further improvement.\comment{it is still far from satisfying. Recently, Generative adversarial network (GAN) has shown great potential for generating or transforming images of exceptional visual fidelity.} In paticular, the current solutions usually ignore the perceptual information of images, which we argue that it benefits the output of a high-quality image while preserving the identity information, especially in facial attributes learning area.
To this end, we propose to train GAN iteratively via regularizing the min-max process with an integrated loss, which includes not only the per-pixel loss but also the perceptual loss. In contrast to the existing methods only deal with either image generation or transformation, our proposed iterative architecture can achieve both of them. Experiments on the multi-label facial dataset CelebA demonstrate that the proposed model has excellent performance on recognizing multiple attributes, generating a high-quality image, and transforming image with controllable attributes.
\keywords{image transformation\and image generation\and perceptual loss\and GAN}
\end{abstract}

\section{Introduction}
\label{intro}
Image generation \cite{gregor2015draw,Wang2017Tag,Odena2017Conditional} and image transformation \cite{Isola2016Image,zhu2016generative,yoo2016pixel-level,Zhu2017Unpaired} are two important topics in computer vision. 
A popular way of image generation is to learn a complex function that maps a latent vector onto a generated realistic image. By contrast, image transformation refers to translating a given image into a new image with modifications on desired attributes or style. Both of them have wide applications in practice. For example, facial composite, which is a graphical reconstruction of an eyewitness's memory of a face~\cite{mcquistonsurrett2006use}, can assist police to identify a suspect. In most situations, police need to search a suspect with only one picture of the front view. To improve the success rate, it is very necessary to generate more pictures of the target person with different poses or expressions. Therefore, face generation and transformation have been extensively studied.

Benefiting from the successes of the deep learning, image generation and transformation have seen significant advances in recent years~\cite{dong2014learning,denoord2016conditional}. With deep architectures, image generation or transformation can be modeled in more flexible ways than traditional approaches. For example, the conditional PixelCNN~\cite{denoord2016conditional} was developed to generate an image based on the PixelCNN. The generation process of this model can be conditioned on visible tags or latent codes from other networks. However, the quality of generated images and convergence speed need improvement\footnote{https://github.com/openai/pixel-cnn}. In \cite{gregor2015draw} and~\cite{yan2015attribute2image}, the Variational Auto-encoders (VAE) \cite{kingma2014auto-encoding} was proposed to generate natural images. 
Recently, Generative adversarial networks (GAN)~\cite{goodfellow2014generative} has been utilized to generate natural images~\cite{denton2015deep} or transform images~\cite{zhu2016generative,Zhu2017Unpaired,zhang2017age,ShuYHSSS17} with conditional settings~\cite{Mirza2014Conditional}.


The existing approaches can be applied to face generation or face transformation respectively, however, there are several disadvantages of doing so. 
First, face generation and face transformation are closely connected with a joint distribution of facial attributes while the current models are usually proposed to achieve them separately (face generation \cite{yan2015attribute2image,li2017generate} or transformation \cite{zhang2017age,ShuYHSSS17}), which may limit the prediction performance.
Second, learning facial attributes has been ignored by existing methods of face generation and transformation, which might deteriorate the quality of facial images.
Third, most of the existing conditional deep models did not consider to preserve the facial identity during the face transformation ~\cite{Isola2016Image} or generation~\cite{yan2015attribute2image}; 


To this end, we propose an iterative GAN with an auxiliary classifier in this paper, which can not only generate high fidelity face images with controlled input attributes, but also integrate face generation and transformation by learning a joint distribution of facial attributes. We argue that the strong coupling between face generation and transformation should benefit each other. And the iterative GAN can learn and even manipulate multiple facial attributes, which not only help to improve the image quality but also satisfy the practical need of editing facial attributes at the same time.
In addition, in order to preserve the facial identity, we regularize the iterative GAN by the perceptual loss in addition to the pixel loss. A quantity metric was proposed to measure the face identity in this paper.


To train the proposed model, we adopt a two-stage approach as shown in Figure~\ref{fig:model}. In the first stage, we train a \emph{discriminator} $\mD$, a \emph{generator} $\mG$, and a \emph{classifier} $\mC$ by minimizing adversarial losses~\cite{goodfellow2014generative} and the label losses as in~\cite{Odena2017Conditional}. In the second stage, $\mG$ and $\mD/\mC$ are iteratively trained with an integrated loss function, which includes a perceptual component~\cite{johnson2016perceptual} between $\mD$'s hidden layers in stage 1 and stage 2, a latent code loss between the input noise $\vz$ and the output noise $\tilde{\vz}$, and a pixel loss between the input real facial images and their corresponding rebuilt version. 

In the proposed model, the generator $\mG$ not only generates a high-quality facial image according to the input attribute (single or multiple) but also translates an input facial image with desired attribute modifications. The fidelity of output images can be highly preserved due to the iterative optimization of the proposed integrated loss.
To evaluate our model, we design experiments from three perspectives, including the necessity of the integrated loss, the quality of generated natural face images with specified attributes, and the performance of face generation. Experiments on the benchmark CelebA dataset~\cite{liu2015faceattributes} have indicated the  promising performance of the proposed model in face generation and face transformation.

\section{Related Work}
\label{sec:related}
\subsection{Facial attributes recognition}
Object Recognition method has been researched for a long while as an active topic~\cite{Farhadi2009Describing,branson2010visual,nilsback2008automated} especially for human recognition, which takes attributes of a face as the major reference ~\cite{kumar2009attribute,cherniavsky2010semi-supervised}. Such attributes include but not limited to \emph{\textbf{natural looks}} like Arched\_Eyebrows, Big\_Lips, Double\_Chin, Male, etc. Besides, some \emph{\textbf{'artificial�attributes}} also contribute to this identification job, like Glasses, Heavy\_Makeup, Wave\_Hair. Even some \emph{\textbf{expression}} like Smiling, Angry, Sad can be labeled as a kind of facial attributes to improve identification. For example, Devi et al. analyze the complex relationship among these multitudinous attributes to categorize a new image~\cite{parikh2011relative}. The early works on automatic expression recognition can be traced back to the early nineties~\cite{bettadapura2012face}.
Most of them~\cite{kumar2009attribute,berg2013poof:,bourdev2011describing} tried to verify the facial attributes based on the methods of HOG~\cite{dalal2005histograms} and SVM~\cite{cortes1995support-vector}. Recently, the development of the deep learning flourished the expression recognition~\cite{zhang2014panda:} and made a success of performing face attributes classification based on CNN (the Convolutional Neural Networks)~\cite{liu2015faceattributes}.
The authors of ~\cite{liu2015faceattributes} even devised a dataset celebA that includes more than 200k face images (each with 40 attributes labels) to train a large deep neural network.
celebA is widely used in facial attributes researches. In this paper, we test iterative GAN with it. FaceNet~\cite{schroff2015facenet:} is another very important work on face attributes recognition that proposed by Google recently. They map the face images to a Euclidean space, thus the distance between any two images that calculated in the new coordinate system shows how similar they are. The training process is based on the simple heuristic knowledge that face images from the same person are closer to each other than faces from different persons. The FaceNet system provides a compare function to measure the similarity between a pair of images. We prefer to utilize this function to measure the facial identity-preserving in quantity.

\subsection{Conditioned Image Generation}
Image generation is an very popular and classic topic in computer vision. 
The vision community has already taken significant steps on the image generation especially with the development of deep learning. 

The conditional models \cite{Mirza2014Conditional,Kingma2014Semi,denoord2016conditional} enable easier controlling of the image generation process. In \cite{denoord2016conditional}, the authors presented an image generation model based on PixelCNN under conditional control. However, it is still not satisfying on image quality and efficiency of convergence. 
In the last three years, generating images with Variational Auto-encoders (VAE) \cite{kingma2014auto-encoding} and GAN \cite{goodfellow2014generative} have been investigated. 
A recurrent VAE was proposed in \cite{gregor2015draw} to model every stage of picture generation. Yan et al. \cite{yan2015attribute2image} used two attributes conditioned VAEs to capture foreground and background of a facial image, respectively.
In addition, the sequential model~\cite{gregor2015draw,denton2015deep} has attracted lots of attention recently. The recurrent VAE ~\cite{gregor2015draw} mimic the process of human drawing but it can only be applied to low-resolution images; in \cite{denton2015deep}, a cascade Laplacian pyramid model was proposed to generate an image from low resolution to a final full resolution gradually. 

GAN \cite{goodfellow2014generative} has been applied to lots of works of image generation with conditional setting since Mirza and Osindero \cite{Mirza2014Conditional} and Gauthier \cite{gauthier14}. 
In \cite{denton2015deep}, a Laplacian pyramid framework was adopted to generate a natural image with each level of the pyramid trained with GAN. In \cite{wang2016generative}, the S$^{2}$GAN was proposed to divide the image generation process into structure and style generation steps corresponding to two dependent GAN.
The Auxiliary classifier GAN (ACGAN) in \cite{Odena2017Conditional} tries to regularize traditional conditional GAN with label consistency. To be more easy to control the conditional tag, the ACGAN extended the original conditional GAN with an auxiliary classifier $\mC$.

\subsection{Image Transformation}
Image transformation is a well-established area in computer vision which could be concluded as ``where a system receives some input image and transforms it into an output image~\cite{johnson2016perceptual}". According to this generalized concept, a large amount of image processing tasks belong to this area such as images denoise~\cite{xie2012image}, generate images from blueprint or outline sketch~\cite{Isola2016Image}, alternate a common picture to an artwork maybe with the style of Vincent van Gogh or Claude Monet~\cite{Zhu2017Unpaired,gatys2015a}, image inpainting~\cite{xie2012image}, the last but not the least, change the appointed facial attributes of a person in the image. 

Most works of image transformation are based on pixel-to-pixel loss. Like us, ~\cite{gatys2015a} transforms images from one style to another with an integrated loss (feature loss and style reconstruction loss) through CNN. Though the result is inspiring, this model is very cost on extracting features on pre-trained model. Recently, with the rapid progress of generative adversarial nets, the quality of output transformed images get better. The existing models of image transforming based on GAN fall into two groups. 
Manipulating images over the natural image manifold with conditional GAN belongs to the first category.
In \cite{zhu2016generative}, the authors defined user-controlled operations to allow visual image editing. The source image is arbitrary, especially could lie on a low-dimensional manifold of image attributes, such as the color or shape.
In a similar way, Zhang et al. \cite{zhang2017age} assume that the age attribute of face images lies on a high-dimensional manifold. By stepping along the manifold, this model can obtain face images with different age attributes from youth to old. The remaining works consist of the second group.
In \cite{Isola2016Image}, the transformation model was built on the conditional setting by regularizing the traditional conditional GAN \cite{Mirza2014Conditional} model with an image mapping loss. Some work considers deploying GAN models for each of the related image domains (.e.g. input domain and output domain), more than one set of adversarial nets then cooperate and restrict with each other to generate high-quality results  \cite{yoo2016pixel-level,Zhu2017Unpaired}. In \cite{yoo2016pixel-level}, a set of aligned image pairs are required to transfer the source image of a dressed person to product photo model. In contrast, the cycle GAN \cite{Zhu2017Unpaired} can learn to translate an image from a source domain to a target domain in the absence of paired samples. This is a very meaningful advance because the paired training data will not be available in many scenarios.

\subsection{Perceptual Losses}

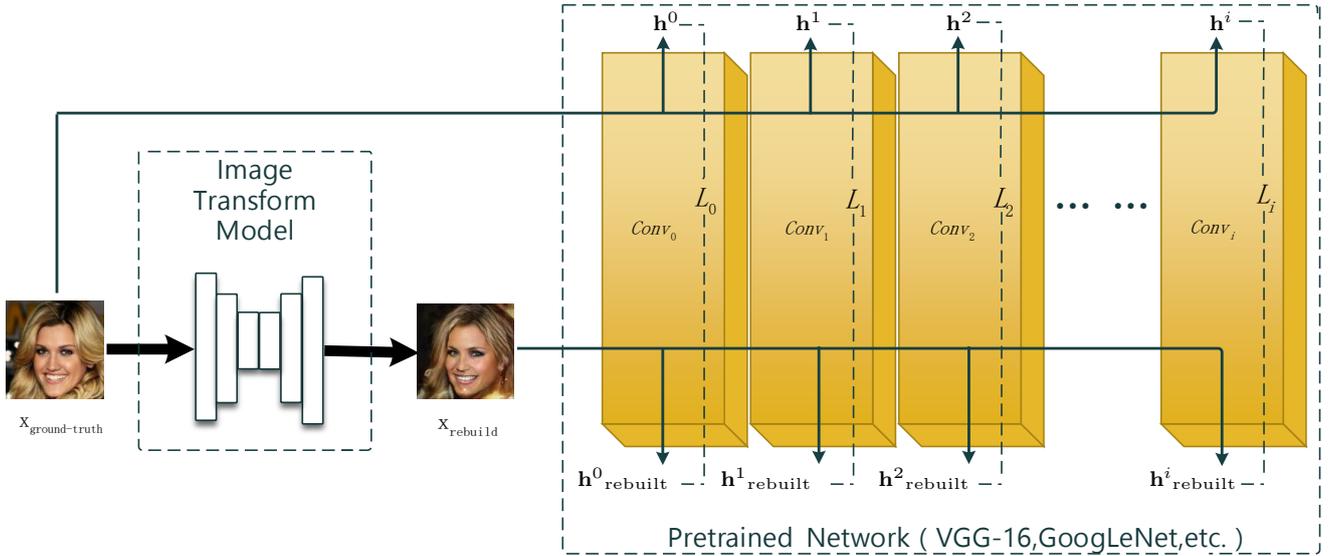
\begin{figure*} [htbp]
\centering
\begin{overpic}[scale=0.6]{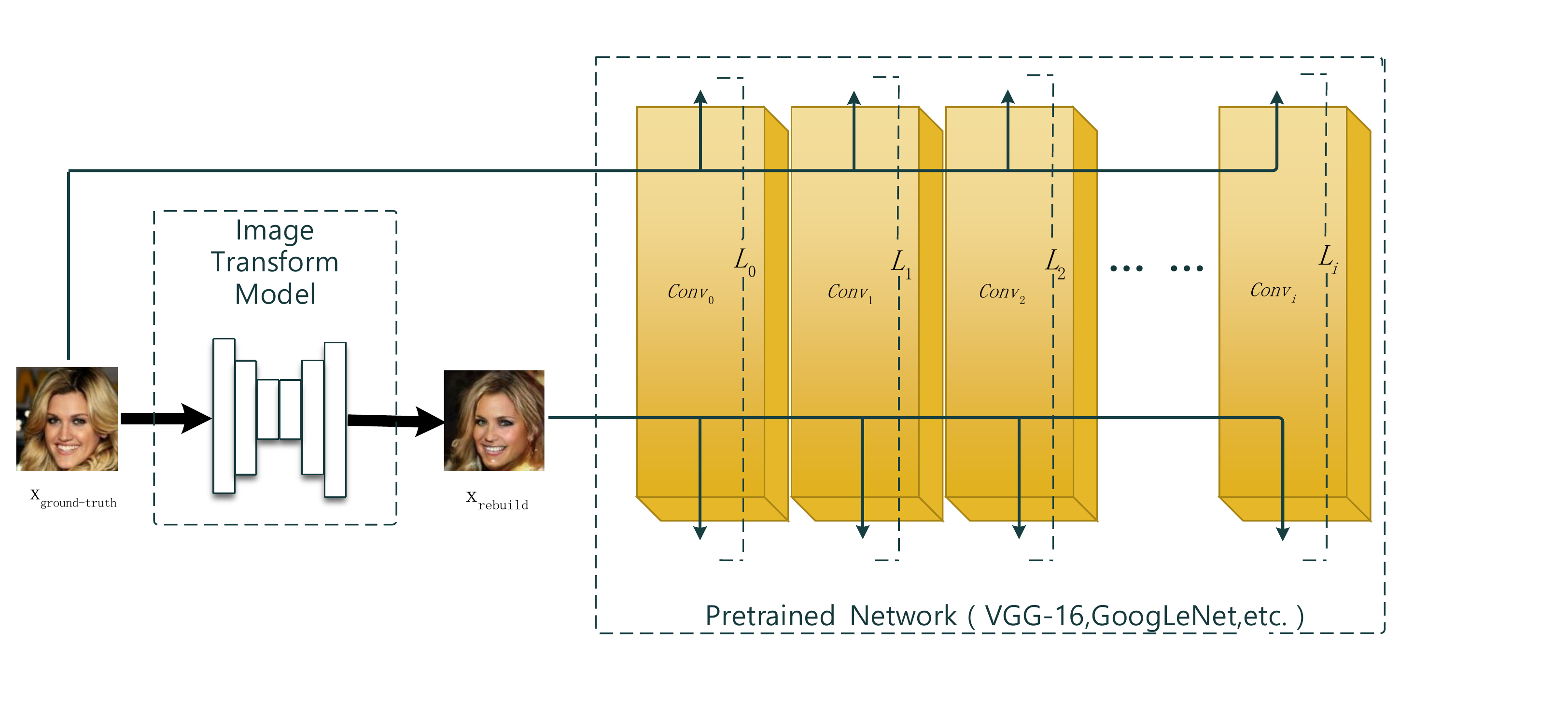}

\put(44,39.5){{\footnotesize $\vh^0$}}
\put(53.5,39.5){{\footnotesize $\vh^1$}}
\put(63.5,39.5){{\footnotesize $\vh^2$}}
\put(81,39.5){{\footnotesize $\vh^i$}}

\put(39,9){{\footnotesize $\vh{^0}_{\text{rebuilt}}$}}
\put(48.5,9){{\footnotesize $\vh{^1}_{\text{rebuilt}}$}}
\put(59,9){{\footnotesize $\vh{^2}_{\text{rebuilt}}$}}
\put(77,9){{\footnotesize $\vh{^i}_{\text{rebuilt}}$}}

\end{overpic}
\caption{Demonstration of how to train an image transformation model~\cite{johnson2016perceptual} with the perceptual loss. A pretrained network (right yellow part) plays a role in extracting feature information of both ground-truth and rebuilt image through several convolutional layers. For each layer, the semantic differences $L_i$ between the two images is calculated. Finally, by reducing the total differences collected through all layers, the image transform model (left part in dashed box) is optimized to rebuild the ground-truth image. 
}
\label{fig:perceptual_loss}
\end{figure*}
A traditional way to minimize the inconsistency between two images is optimizing the pixel-wise loss~\cite{tatarchenko2016multi-view,Isola2016Image,zhang2016colorful}. However, the pixel-wise loss is inadequate for measuring the variations of images, since the difference computed in pixel element space is not able to reflect the otherness in visual perspective. We could easily and elaborately produce two images which are absolutely different observed in human eyes but with minimal pixel-wise loss and vice verse. Moreover, using single pixel-wise loss often tend to generate blurrier results with the ignorance of visual perception information~\cite{Isola2016Image,Larsen2015Autoencoding}.
In contrast, the perceptual loss explores the
discrepancy between high-dimensional representations of images extracted from a well-trained CNN (for example VGG-16~\cite{simonyan2015very}) can overcome this problem. By narrowing the discrepancy between ground-truth image and output image from a high-feature-level perspective, the main visual information is well-reserved after the transformation.  
Recently, the perceptual loss has attracted much attention in image transformation area  ~\cite{johnson2016perceptual,gatys2015a,Wang2017Perceptual,dosovitskiy2016generating}. By employing the perceptual information,~\cite{gatys2015a} perfectly created the artistic images of high perceptual quality that approaching the capabilities of human;~\cite{ledig2016photo-realistic} highlighted the applying of perceptual loss to generate super-resolution images with GAN, the result is amazing and state-of-art;~\cite{Wang2017Perceptual} proposed a generalized model called perceptual adversarial networks (PAN) which could transform images from sketch to colored, semantic labels to ground-truth, rainy day to de-rainy day etc. Zhu et al.~\cite{zhu2016generative} focused on designing a user-controlled or visual way of image manipulation utilize the perceptual similarity between images that defined over a low dimensional manifold to visually edit images. 

A convenient way of calculating the perceptual loss between a ground-truth face $\vx_{\text{ground-truth}}$ and a transformed face $\vx_{\text{rebuilt}}$ is to input them to a Convolutional Neural Networks (such as the VGG-16~\cite{simonyan2015very} or the GoogLeNet~\cite{szegedy2015going} ) then sum the differences of their representations from each hidden layers~\cite{johnson2016perceptual} of the network.
As shown in Fig.~\ref{fig:perceptual_loss}~\cite{johnson2016perceptual}, both of the $\vx_{\text{ground-truth}}$ and $vx_{\text{rebuilt}}$ was passed to the pretrained network as inputs. 
The differences between their corresponding latent feature matrices $\vh^{i}$ and $\vh^{i}_{\text{rebuilt}}$ can be calculated ($L_i= \|\vh_{\text{rebuilt}}^{i} - \vh^{i}\|$).

The final perceptual loss can be represented as follows,
\begin{gather*}\label{eq:original_perceptual_loss}
L_{per} = \mathbb{E}_{\{\vh_{\text{rebuilt}}^{i},\vh^{i}\}}[\sum_{i=1}\alpha_{i}\|\vh_{\text{rebuilt}}^{i} - \vh^{i}\|]
\end{gather*}
Where $\vh_{\text{rebuilt}}^{i},\vh^{i}$ are the feature matrices output from each convolutional layers, $\alpha_i$ are the parameters to balance the affections from each layers.

For facial image transformation, perceptual information shows its superiority in preserving the facial identity. To take advantages of this property of the perceptual loss, in our proposed model, we leverage it to keep the consistency of personal identity between two face images. In particular, we choose to replace the popular pretrained Network with the discriminator network to decrease the complexity. We will discuss this issue in detail later.


\section{Proposed Model}
\label{sec:model}
We first describe the proposed model with the integrated loss, then explain each component of the integrated loss.

\begin{figure*}[htbp]
  \centering
  \includegraphics[width=14cm,height=8cm]{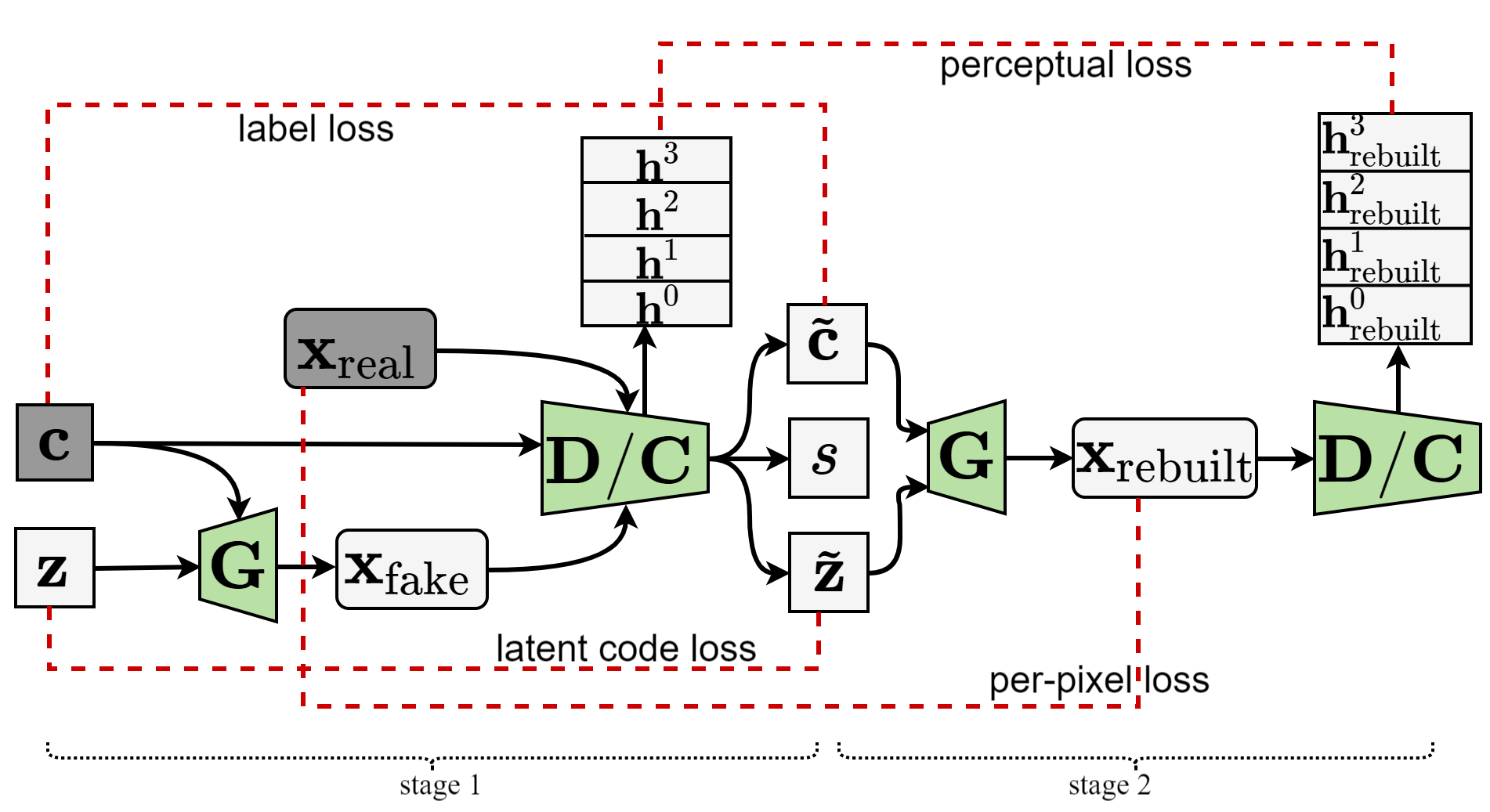}
  \caption{The architecture of our proposed model. Training: the model begin with a original GAN process, the generator $\mG$ takes a random noise vector $\vz$ and label $\vc$ as inputs. It outputs a generated face image $\vx_{\text{fake}}$. The discriminator $\mD$ receives both $\vx_{\text{fake}}$ and a real image $\vx_{\text{real}}$, and outputs the probability distribution over possible image sources. An auxiliary classifier $\mC$, which shares layers (without the last layer) with $\mD$, predicts a label $\tilde{\vc}$ and outputs reconstructed noise code $\tilde{\vz}$. Meanwhile, $\mD$ outputs $s$ (indicates whether the image is fake or real). At the second stage, $\mG$ rebuilds the $\vx_{\text{real}}$ by generating a $\vx_{\text{rebuilt}}$ with $\tilde{\vz}$ and $\tilde{\vc}$. Then $\mD$ iteratively receives $\vx_{\text{rebuilt}}$ and updates the hidden layers (from $\vh$ to $\vh_{\text{rebuilt}}$).
  During testing, we can deal with both face generation and transformation.
  By sampling a random vector $\vz$ and a desired facial attribute description, a new face can be generated by $\mG$. For face transformation, we feed the discriminator $\mD$ a real image $\vx_{\text{real}}$ together with its attribute labels $\vc$, then we get a noise representation $\tilde{\vz}$ and label vector $\tilde{\vc}$. The original image can be reconstructed with $\tilde{\vz}$ and $\tilde{\vc}$ by feeding them back to the generator $\mG$. The new image generated by $\mG$ is named $\vx_{\text{rebuilt}}$. Alternatively, we can modify the content of $\tilde{\vc}$ before inputting $\mG$. According to the modification of labels, the corresponding attributes of the reconstructed image will be transformed.
  }\label{fig:model}
\end{figure*}

\subsection{Problem Overview}
To our knowledge, most of the existed methods related to image processing are introduced for one single goal, such as facial attributes recognition, generation or transformation. Our main purpose is to develop a multi-function model that is capable of managing these tasks altogether, end-to-end. 
\begin{itemize}
    \item{\emph{\textbf{Facial attributes recognition}}
    
    By feeding a source image $\vx$ (it doesn't matter if it is real or fake) to the $\mD$, the classifier $\mC$ would output the probability of concerned facial attributes of $\vx$.}
    \item{\emph{\textbf{Face Generation} }
    
    By sampling a random vector $\vz$ and a desired facial attributes description, a new face can be generated by $\mG$.}
    \item{\emph{\textbf{Face Transformation}}
    
    For face transformation, we feed the discriminator $\mD$ a real image $\vx_{\text{real}}$, then we get a noise representation $\tilde{\vz}$ and label vector $\tilde{\vc}$. The original image can be reconstructed with $\tilde{\vz}$ and $\tilde{\vc}$ by feeding them back to the generator $\mG$. The new image generated by $\mG$ is named $\vx_{\text{rebuilt}}$. Alternatively, we can modify the content of $\tilde{\vc}$ recognized by classifier $\mC$ before we input it into $\mG$. According to the modification of labels, the corresponding attributes of the reconstructed image then  transformed.}
\end{itemize}
To this end, We design a variant of GAN with iterative training pipeline as shown in  Fig. \ref{fig:model}, which is regularized by a combination of loss functions, each of which has its own essential purpose. 

In specific, the proposed iterative GAN includes a generator $\mG$, a discriminator $\mD$, and a classifier $\mC$. 
The discriminator $\mD$ and generator $\mG$ along with classifier $\mC$ can be re-tuned with three kinds of specialized losses. In particular, the perceptual losses captured by the hidden layers of discriminator will be back propagated to feed the network with the semantic information. In addition, the difference between the noise code detached by the classifier $\mC$ and the original noise will further tune the networks. Last but not least, we let the generator $\mG$ to rebuild the image $\vx_{\text{rebuilt}}$, then the error between $\vx_{\text{rebuilt}}$ and the original image $\vx_{\text{real}}$ will also feedback. We believe this iterative training process learns the facial attributes Well and the experiments demonstrate that aforementioned three loss functions play indispensable roles in generating natural images.

The object of iterative GAN is to obtain the optimal $\mG$, $\mD$, and $\mC$ by solving the following optimization problem,
\begin{equation}\label{eq:totalLoss}
\argmin \limits_{\mG} \max \limits_{\mD,\mC} L_{\text{ACGAN}}(\mG,\mD,\mC) + L_{\text{inte}}(\mG,\mD,\mC)
\end{equation}
where $L_{\text{ACGAN}}$ is the ACGAN \cite{Odena2017Conditional} loss term to guarantee the excellent ability of classification, $L_{\text{inte}}$ is the integrated loss term which contributes to the maintenance of source image's features. We appoint no balance parameters for the two loss terms because $L_{\text{inte}}$ is a conic combination of another three losses (we will introduce them later). The following part will introduce the definitions of the two losses in detail.

\subsection{ACGAN Loss}
\label{sec:acganLoss}
To produce higher quality images, the ACGAN \cite{Odena2017Conditional} extended the original adversarial loss~\cite{goodfellow2014generative} to a combination of the adversarial and a label consistency loss. Thus, the label attributes are learned during the adversarial training process. The label consistency loss makes a more strict constraint over the training of the min-max process, which results in producing higher quality samples of the generating or transforming. Inspired by the ACGAN, we link the classifier $\mC$ into the proposed model as an auxiliary decoder. But for the purpose of recognizing multi-labels, we modify the representation of output labels into an $n$-dimensional vector, where $n$ is the number of concerned labels. The ACGAN loss function is as follows,
\begin{equation*}\centering
    L_{\text{ACGAN}}(\mG,\mD,\mC) =  L_{\text{adv}} + L_{\text{label}}
\end{equation*}
where the $L_{\text{adv}}$ is the min-max loss defined in the original GAN \cite{goodfellow2014generative,Mirza2014Conditional} (Section \ref{sec:adversarialloss}), $L_{\text{label}}$ is the label consistency loss \cite{Odena2017Conditional} of classifier $\mC$ (see Section \ref{sec:labelloss}).

\subsubsection{Adversarial Loss}
\label{sec:adversarialloss}
As a generative model, GANs consists of two neural networks \cite{goodfellow2014generative}, the generative network $\mG$ which chases the goal of learning distribution of the real dataset to synthesize fake images and the discriminative network $\mD$ which endeavor to predict the source of input images. The conflicting purposes forced $\mG$ and $\mD$ to play a min-max game, and regulated the balance of the adversarial system. 

In standard GAN training, the generator $\mG$ takes a random noise variable $\vz$ as input and generates a fake image $\vx_{\text{fake}}$ = $\mG(\vz)$. In opposite, the discriminator $\mD$ takes both the synthesized and native images as inputs and predicts the data sources. We follow the form of adversarial loss in ACGAN \cite{Odena2017Conditional}, which increases the input of $\mG$ with additional conditioned information $\vc$ (the attributes labels in the proposed model). The generated image hence depends on both the prior noise data $\vz$ and the label information $\vc$, this allows for reasonable flexibility to combine the representation, $\vx_{\text{fake}} = \mG(\vz,\vc)$. Notice
that unlike the CGANs \cite{Mirza2014Conditional}, in our model, the input of $\mD$ remains in the primary pattern without any conditioning.

During the training process, the discriminator $\mD$ is forced to maximize the likelihood it assigns the correct data source, and the generator $\mG$ performs oppositely to fool the $\mD$ as following,

\begin{equation*}
\begin{aligned}
L_{\text{adv}}(\mG,\mD) &= \mathbb{E}_{\vx, \vc\sim p(\vx, \vc)}[\log \mD(\vx)] \\
&+ \mathbb{E}_{\vz \sim p(\vz)}[\log(1-\mD( \mG( \vz|\vc)))]
\end{aligned}
\end{equation*}

\subsubsection{The Consistency of Data Labels}
\label{sec:labelloss}
The label loss function of the classifier $\mC$ is as following,
\begin{equation*}
    L_{\text{label}} = \mathbb{E}_{\vx \in \{\vx_{\text{real}}, \vx_{\text{fake}}\}}[\log \mC(\mD(\vx))]
\end{equation*}

Either for the task of customized images generation or appointed attributes transformation, proper label prediction is necessary to resolve the probability distribution over the attributes of samples. We take the successful experiences of ACGAN~\cite{Odena2017Conditional} for reference to keep the consistency of data labels for each real or generated sample.
During the training process, the real images $\vx_{\text{real}}$ as well as the fake ones $\vx_{\text{fake}} = \mG(\vz,\vc)$ are all fed to the discriminator $\mD$, the share layer output from $\mD$ then is passed to classifier $\mC$ to get the predicted labels $\tilde{\vc} = \log(\vc - \mC(\mD(\vx)))$. The loss of this predicted labels $\tilde{\vc}$ and the actual labels $\vc$ then is propagated back to optimize the $\mG$, $\mD$, and $\mC$.

\subsection{Integrated Loss }


The ACGAN loss in Equation (\ref{eq:totalLoss}) keeps the generated images by the iterative GAN lively. Additionally, to rebuild the information of the input image, we introduce the integrated loss that combines the per-pixel loss, the perceptual loss, and the latent code loss with parameter $\bm{\lambda} = (\lambda_1,\lambda_2,\lambda_3)\in \RR^3$,
\begin{equation*}
    L_{\text{inte}} = \lambda_1 L_{\text{per}} + \lambda_2 L_{\text{pix}} + \lambda_3 L_{\vz}, \quad \lambda_i \geq 0.
\end{equation*}
The conic coefficients $\bm{\lambda}$ also suggests that we do not need to set a trade-off parameter in Eq. (\ref{eq:totalLoss}). 
We study the necessity of the three components by reconstruction experiments as shown in Section \ref{sec:reconstrction}.
These experiments suggest that combining three loss terms together instead of using only one of them clearly strengthens the training process and improves the quality of reconstructed face image.
During the whole training process, we set $\lambda_1 = 2.0$, $\lambda_2 = 0.5$ and $\lambda_3 = 1.0$. 

We then introduce three components of the integrated loss as following.

\subsubsection{Per-pixel Loss}
\label{sec:pixelLoss}


Per-pixel loss \cite{tatarchenko2016multi-view,dong2014learning} is a straightforward way to pixel-wisely measure the difference between two images, the input face $\vx_{\text{real}}$ and the rebuilt face $\vx_{\text{rebuilt}}$ as follows,
\begin{equation*}
    L_{\text{pix}} = \mathbb{E}[\|\vx_{\text{real}}-\vx_{\text{rebuilt}}\|] 
\end{equation*}
where
$\vx_{\text{rebuilt}} =
\mG(\tilde{\vz},\tilde{\vc})$
is the generator $\mG$ that reconstructs the real image $\vx_{\text{real}}$ based on predicted values $\tilde{\vz}$ and $\tilde{\vc}$. The per-pixel loss forces the source image and destination image as closer as possible within the pixel space. Though it may fail to capture the semantic information of an image (a tiny and invisible diversity in human-eyes may lead to huge per-pixel loss, vice versa), we still think it is a very important measure for image reconstruction that should not be ignored.

The process of rebuilding $\tilde{\vz}$ and $\tilde{\vc}$ is demonstrated in Fig. \ref{fig:model}. Given a real image $\vx_{\text{real}}$, the discriminator $\mD$ extracts a hidden map $\vh^3_{\text{real}}$ with its 4$th$ convolution layer. 
Then $\vh^3_{\text{real}}$ is linked to two different full connected layers and they output a 1024-dimension share layer (a layer shared with C) and a scalar (the data source indicator s) respectively. 
The classifier $\mC$ also has two full-connected layers.
It receives the 1024-dimension share layer from $\mD$ as an input and outputs the rebuilt noise $\tilde{\vz}$ and the predicted label $\tilde{\vc}$ with its two full-connected layers as shown in Fig. \ref{fig:architecture of model}:
\begin{equation*}
\tilde{\vz},\tilde{\vc} = \mC(\mD(\vx_{\text{real}})).
\end{equation*}

\subsubsection{Perceptual Loss}
\label{sec:perceptualLoss}

Traditionally, per-pixel loss \cite{kingma2014auto-encoding,tatarchenko2016multi-view,dong2014learning}is very efficient and popular in reconstructing image. However, the pixel-based loss is not a robust measure since it cannot capture the semantic difference between two images~\cite{johnson2016perceptual}. For example, some unignorable defects, such as blurred results (lack of high-frequency information), artifacts (lack of perceptual information)~\cite{Wang2017Perceptual}, often exist in the output images reconstructed via the per-pixel loss.
To balance these side effects of the per-pixel loss, we feed the training process with the perceptual loss  \cite{johnson2016perceptual,dosovitskiy2016generating,Wang2017Perceptual} between $\vx_{\text{real}}$ and $\vx_{\text{rebuilt}}$. We argue that this perceptual loss captures the discrepancy between two images in semantic space. 

To reduce the model complexity, we calculate the perceptual loss on convolutional layers of the discriminator $\mD$ rather than on third-part pre-trained networks like VGG, GoogLeNet.
Let $\vh^i$ and $\vh^i_{\text{rebuilt}}$ be the feature maps extracted from the $i$-th layer of $\mD$ (with real image and rebuilt image respectively), then the perceptual loss $L_{\text{per}}$ between the original image and the rebuilt one is defined as follows:
\comment{
\begin{equation}
    L_{\text{per}} = \mathbb{E}[\log P(\vh^i_{\text{rebuilt}}|\vh^i)]
\end{equation}
}
\begin{equation*}
    L_{\text{per}} = \mathbb{E}[\|\vh^i_{\text{rebuilt}}-\vh^i\|].
\end{equation*}
The minimization of the $L_{\text{per}}$ forces the perceptual information in the rebuilt face  to be consistent with the original face.

\subsubsection{Latent Code Loss}
\label{sec:zLoss}
The intuitive idea to rebuild the source images is that we assume that the latent code of face attributes lie on a manifold $\mathcal{M}$ and faces can be generated by sampling the latent code on different directions along the $\mathcal{M}$. 

In the train process, the generator $\mG$ takes a random latent code $\vz$ and label $\vc$ as an input and outputs the fake face $\vx_{\text{fake}}$. Then the min-max game forces the discriminator $\mD$ to discriminate between $\vx_{\text{fake}}$ and the real image $\vx_{\text{real}}$.
Meanwhile, the auxiliary classifier $\mC$, which shares layers (without the last layer) with $\mD$, detach a reconstructed latent code $\tilde{\vz}$. At the end of the min-max game, both $\vz$ and $\tilde{\vz}$ should share a same location on the $\mathcal{M}$ because they are extracted from a same image. Hence, we construct a loss $\tilde{\vz}$ and the random $\vz$ to regularize the process of image generation, i.e.,


\comment{\begin{equation}
    L_{\vz} = \mathbb{E}[\log P(\tilde{\vz}|\vz)]
\end{equation}}

\begin{equation*}
    L_{\vz} = \mathbb{E}[\|\tilde{\vz}-\vz\|]
\end{equation*}

In this way, the latent code $\tilde{\vz}$ detached by the classifier $\mC$ will be aligned with $\vz$.

\section{Network Architecture}

\begin{figure*}[htbp]
  \centering
  \includegraphics[scale= 0.4]{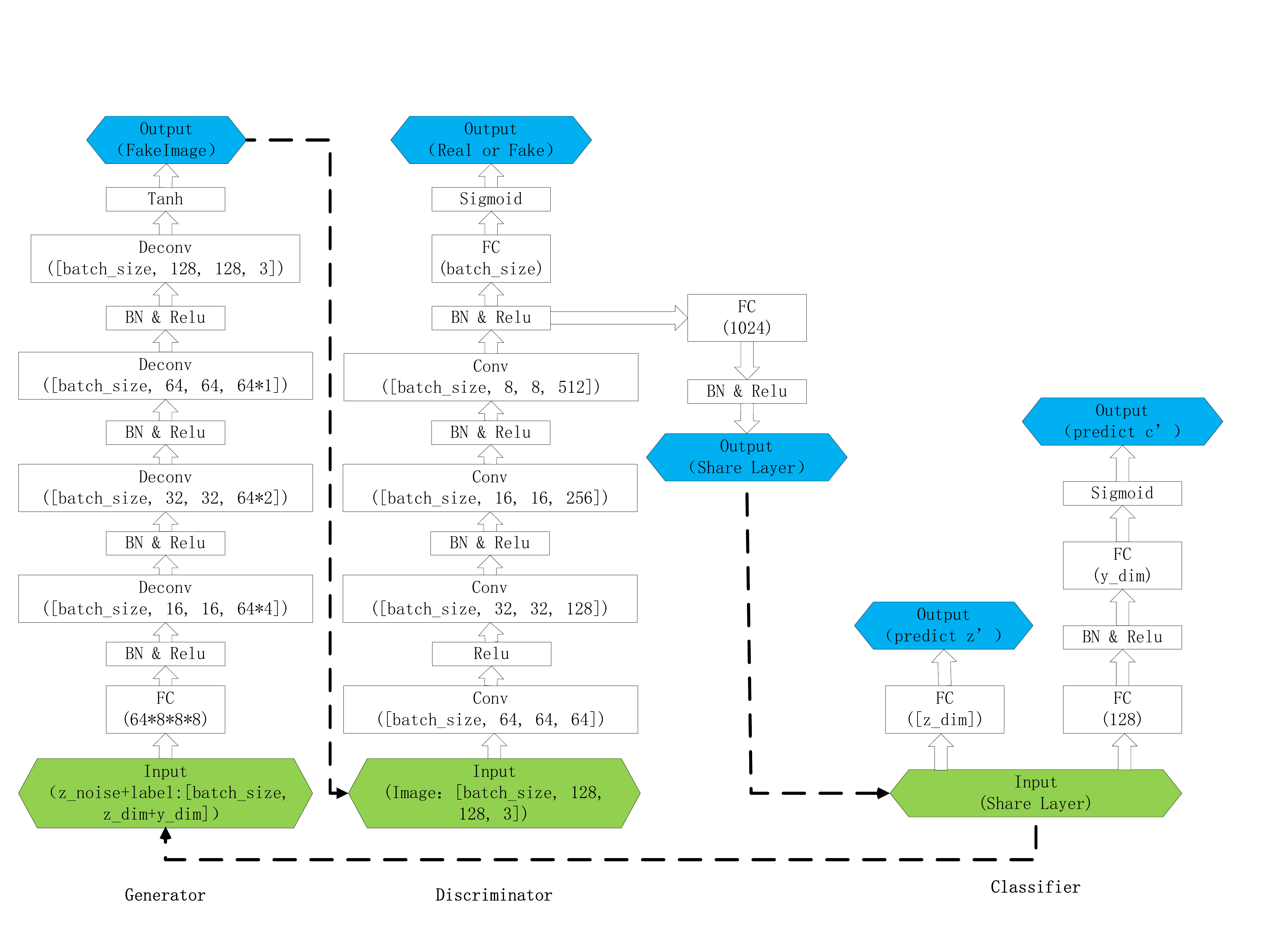}\\
  \caption{The overview of the network architecture of iterative GAN.}\label{fig:architecture of model}
\end{figure*}
Iterative GAN includes three neural networks. The generator consists a fully-connected layer with 8584 neurons, four de-convolutional layers and each has 256, 128, 64 and 3 channels with filer size 5*5. All filter strides of the generator $\mG$ are set as 2*2. After processed by the first fully-connected layer, the input noise $\vz$ with 100 dimensions is projected and reshaped to [Batch\_size, 5, 5, 512], the following 4 deconvolutional layers then transpose the tensor to  [Batch\_size, 16, 16, 256], [Batch\_size, 32, 32, 128], [Batch\_size, 64, 64, 64], [Batch\_size, 128, 128, 3], respectively. The tensor output from the last layer is activated by the \emph{tanh} function.

The discriminator is organized almost in the reverse way of the generator. It includes 4 convolutional layers with filter size 5*5 and stride size 2*2. The training images with shape [Batch\_size, 128, 128, 3] then transpose through the following 4 convolutional layers and receive a tensor with shape [Batch\_size, 5, 5, 512]. The discriminator needs to output two results, a shared layer for the classifier and probability that indicates if the input image is a fake one. We flat the tensor and pass it to classifier as input for the former purpose. To get the reality of image, we make an extra full connect layer which output [Batch\_size, 1] shape tensor with \emph{sigmoid} active function.

The classifier receives the shared layer from the discriminator which contains enough feature information. 
We build two full-connected layers in the classifier, one for the predicted noise  $\tilde{\vz}$ and the other for the output predicted labels  $\tilde{\vc}$. 
And we use the \emph{tanh} and  \emph{sigmoid} function to squeeze the value  $\tilde{\vz}$ to (-1,1),  $\tilde{\vc}$ to (0, 1) respectively. And the Fig. \ref{fig:architecture of model} shows how we organize the networks of iterative GAN.
\begin{table}
\centering
\caption{Analysis of time consumption for training.}
\label{tab:timeconsumption1}       
\begin{tabular}{ccccc}
\hline\noalign{\smallskip}
&epoch & image Size  & cost time & \\
\noalign{\smallskip}\hline\noalign{\smallskip}
\multirow{3}{*}{Training }&10 &128*128 & 106,230 sec & \\
    &20 &128*128 & 23,9460 sec &\\
	&50 &128*128 & 600,030 sec &\\
\noalign{\smallskip}\hline
\end{tabular}
\end{table}

\begin{table*}
\centering
\caption{Analysis of time consumption for testing.}
\label{tab:timeconsumption2}       
\begin{tabular}{lcccc}
\hline\noalign{\smallskip}
&image num & input image size & output image size & cost time  \\
\noalign{\smallskip}\hline\noalign{\smallskip}
Rebuild &64 & 128*128  & 128*128 &3.2256 sec \\
Generate &64 & $\backslash$  & 128*128 & 2.1558 sec \\
\noalign{\smallskip}\hline
\end{tabular}
\end{table*}

\subsection{Optimization}
Following the DCGAN~\cite{radford2016unsupervised}, we adopt the Adam optimizer~\cite{adam2015ICLR} to train proposed iterative GAN. The learning rate $\alpha$ is set to 0.0002 and $\beta_1$ is set to 0.5, $\beta_2$ is 0.999 (same settings as DCGAN). To avoid the imbalanced optimization between the 2 competitors G and D, which happened commonly during GANs's training and caused the vanishment of gradients, we set 2 more parameters to control the update times of D and G in each iteration. While the loss of D is overwhelmingly higher than G, we increase the update times of D in this iteration, and vice versa. By using this trick, the stability of training process is apparently approved.

\subsection{Statistics for Training and Testing}
We introduce the time costs of training and testing in this section.
The proposed iterative GAN model was trained on an Intel Core i5-7500 CPU@3.4GH$_z$ with 4 cores and NVIDIA 1070Ti GPU. The training process goes through 50 epochs and shows the final results. The Analysis of the time consumption (including both training and forward propagation process) are shown in Table \ref{tab:timeconsumption1} and Table \ref{tab:timeconsumption2} respectively.

\section{Experiment}
\label{sec:experiments}
We perform our experiment for multiple tasks to verify the capability of iterative GAN model: recognition of facial attributes, face images reconstruction, face transformation, and face generation with controllable attributes. 
\subsection{Dataset}
We run the iterative GAN model on celebA dataset \cite{liu2015faceattributes} which is based on celebFace+~\cite{sun2014deep}. celebA is a large-scale face image dataset contains more than 200k samples of celebrities. Each face contains 40 familiar attributes, such as \emph{\textbf{Bags\_Under\_Eyes}}, \emph{\textbf{Bald}}, \emph{\textbf{Bangs}}, \emph{\textbf{Blond\_Hair}}, etc. Owing to the rich annotations per image, celebA has been widely applied to face visual work like face attribute recognition, face detection, and landmark (or facial part) localization. We take advantage of the rich attribute annotations and train each label in a supervised learning approach.


We split the whole dataset into 2 subsets: 185000 images of them are randomly selected as training data,
the remaining 15000 samples are used to evaluate the results of the experiment as the test set. 

We crop the original images of size 178 * 218 into 178 * 178, then further resize them to 128 * 128 as the input samples. The size of the output (generated) images are as well as the inputs.

\subsection{The Metric of Face Identity}
Given a face image, whatever rebuilding it or transforming it with customized attributes, we have to preserve the similarity (or identity) between the input and output faces. It is very important for human face operation because the maintenance of the primary features belong to the same person is vital during the process of face transformation. Usually, the visual effect of face identity-preserving can only be observed via naked eyes.

Besides visual observation, in this paper, we choose FaceNet~\cite{schroff2015facenet:} to define the diversity between a pair of images. In particular, FaceNet accepts two images as input and output a score which reveal the similarity. A lower score indicates that the two images are more similar, and vice versa. In other words, FaceNet provides us a candidate metric on the face identity evaluation. 
We take it as an important reference in the following related experiments.




\subsection{Results}
\subsubsection{Recognition of facial attributes (multi-labels)} 
Learning face attributes is fundamental to face generation and transformation.
Previous work learned to control single attribute \cite{li2017generate} or multi-category attributes~\cite{Odena2017Conditional} through a \emph{softmax} function for a given input image. However, the natural face images are always associated with multiple labels. To the best of our knowledge, recognizing and controling the multi-label attributes for a given facial image are among most challenging issues in the community. In our framework, the classifier $\mC$ accepts a 1024 dimensions shared vector that outputted by the discriminator $\mD$, then $\mC$ squash it into 128 dimensions by a full connection. To output the multiple labels, we just need to let the classifier $\mC$ to compress the 128 dimension median vector into $d$ dimensions (as shown in Fig. \ref{fig:architecture of model}, $d$ is the dimension of label vector, $d=40$ in this paper). By linking the $d$ dimensions vector to $d$ $sigmoid$ mappings, the classifier $\mC$ output the predictions of attribute labels finally.
\begin{figure} 
\centering
\includegraphics[width=2.2in,height=1.2in]{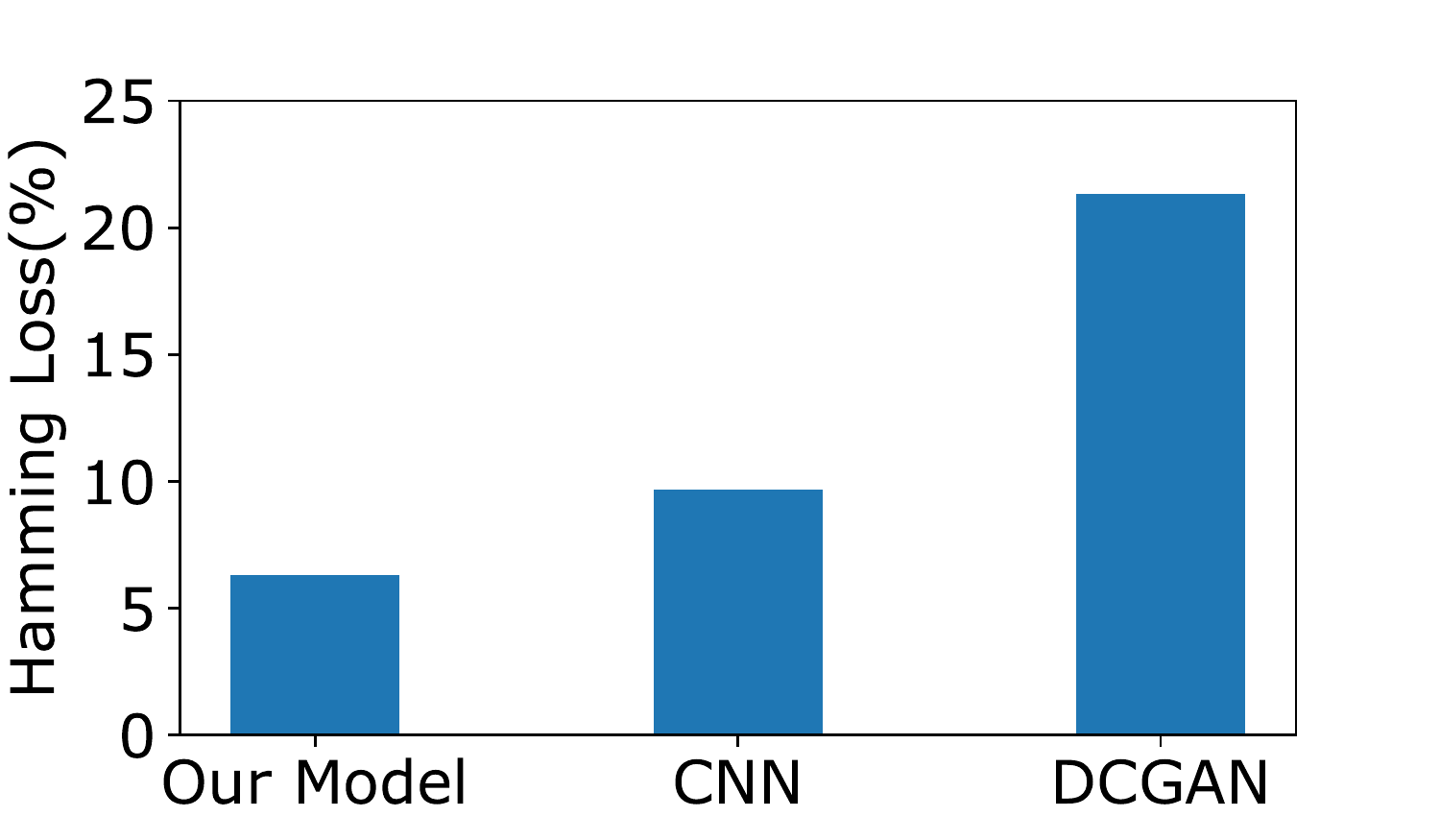}
\caption{The hamming loss of attribute prediction.}
\label{fig:attriPre}
\end{figure}

We feed the images in test set to the classifier $\mC$ and calculate the hamming loss for the multi-label prediction. The statistic of hamming loss is over all the 40 labels associated with each test samples. Two methods are selected as baselines. The first one is native DCGAN with L2-SVM classifier which is reported superior to $k$-means and $k$-NN \cite{radford2016unsupervised}. The other is a Convolutional Neural Network. We train both referenced models on celebA. For the DCGAN, the discriminator extracts the features and feeds it to the linear L2-SVM with Euclidean distance for unsupervised classification. Mean while, the CNN model outputs the predicted labels directly after standard process of convolution training. 

Fig. \ref{fig:attriPre} illustrates the hamming loss of the three algorithms. It is clear to see that the iterative GAN significantly outperforms DCGAN+L2-SVM and CNN. We speculate that the proposed joint architecture on both face generation and transformation regularized by the integrated loss make the facial attribute learning of iterative GAN is much easier than the baselines.

Besides the hamming loss statistics of hamming loss, we also visualize part of the results in Table \ref{tab:attriRecog}.
Row 2, 3 and 4 in Table \ref{tab:attriRecog} illustrate three examples in the test set.  
Row 2 and 3 are the successful cases, while row 4 shows a failed case on predicting \emph{\textbf{HeavyMakeUp}} and \emph{\textbf{Male}}.

\begin{table}
\scalebox{1}{
\begin{tabular}{|c|c|c|c|}
 \hline
Target Image& Attribute & Truth & Prediction\\
 \hline
\multirow{10}{*}{\includegraphics[width = .12\textwidth]{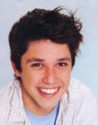}} & Bald & -1 &0.0029 \\
&Bangs & -1 &0.0007 \\
&BlackHair & 1 &0.8225 \\
&BlondeHair & -1 &0.2986 \\
&EyeGlass & -1 &0.0142 \\
&Male & 1 &0.8669 \\
&Nobeard & 1 &0.7255 \\
& Smiling & 1 &0.9526 \\
&WaveHair & 1 &0.6279 \\
&Young & 1 &0.6206 \\
 \hline
 \multirow{7}{*}{\includegraphics[width = .09\textwidth]{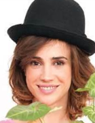}} & Attractive & 1 &0.7758\\
 & Bald & -1 &0.1826\\
 & Male & -1 &0.00269\\
 & Smiling & 1 &0.7352\\
 & HeavyMakeUp & 1 &0.5729\\
  & Wearinghat & 1 &0.7699\\
 & Young & 1 &0.8015\\
\hline
\multirow{5}{*}{\includegraphics[width=.06\textwidth]{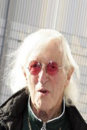}} & Attractive & -1 &0.4629\\
 & Bald & -1 &0.6397\\
 & EyeGlass & 1 &0.8214\\
 & \textbf{HeaveMakeUp} & \textbf{-1} &\textbf{0.7566}\\
 & \textbf{Male} & \textbf{1} &\textbf{0.3547}\\
\hline
\end{tabular}
}
\caption{Demonstrate the example of classification result of iterative GAN. The listed ground-truth tags of target image are expressed by two integer 1 and -1. Row 1 and 2 show the exactly correct prediction examples. Row 3 demonstrates the miss-classification example: the classifier failed to determine the attribute Heavy\_makeup and Male of the face which expressed in black font.}\label{tab:attriRecog}
\end{table}

\subsubsection{Reconstruction}
\label{sec:reconstrction}
In this experiment, we reconstruct given target faces in 4 different settings (per-pixel\_loss,$\vz$\_loss, $\vz$\_loss+per-pixel\_loss, and the integrated Loss) separately.
This experiment proves that the iterative GAN provided with integrated loss has a strong ability to deal with the duties mentioned above and archive the similar or better results than previous works.

By feeding the noise vector $\tilde{\vz}$ and $\tilde{\vc}$, the generator $\mG$ can reconstruct the original input image with its attributes preserved ($\vx_{\text{rebuilt}}$ in Fig. \ref{fig:model}). 
We will evaluate the contributions of integrated loss in this experiment of face reconstruction. 
In detail, we run 4 experiments by regularizing the ACGAN Loss with: only per-pixel loss, latent code loss($\vz$\_loss), latent code loss($\vz$\_loss) + per-pixel loss, and latent code loss($\vz$\_loss) + per-pixel loss + perceptual loss. 

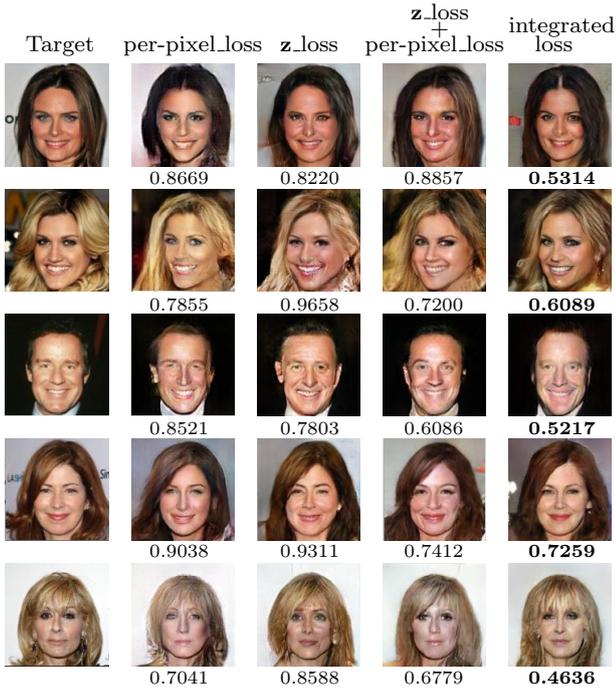
\begin{figure} 
\centering
\begin{overpic}[scale=0.4]{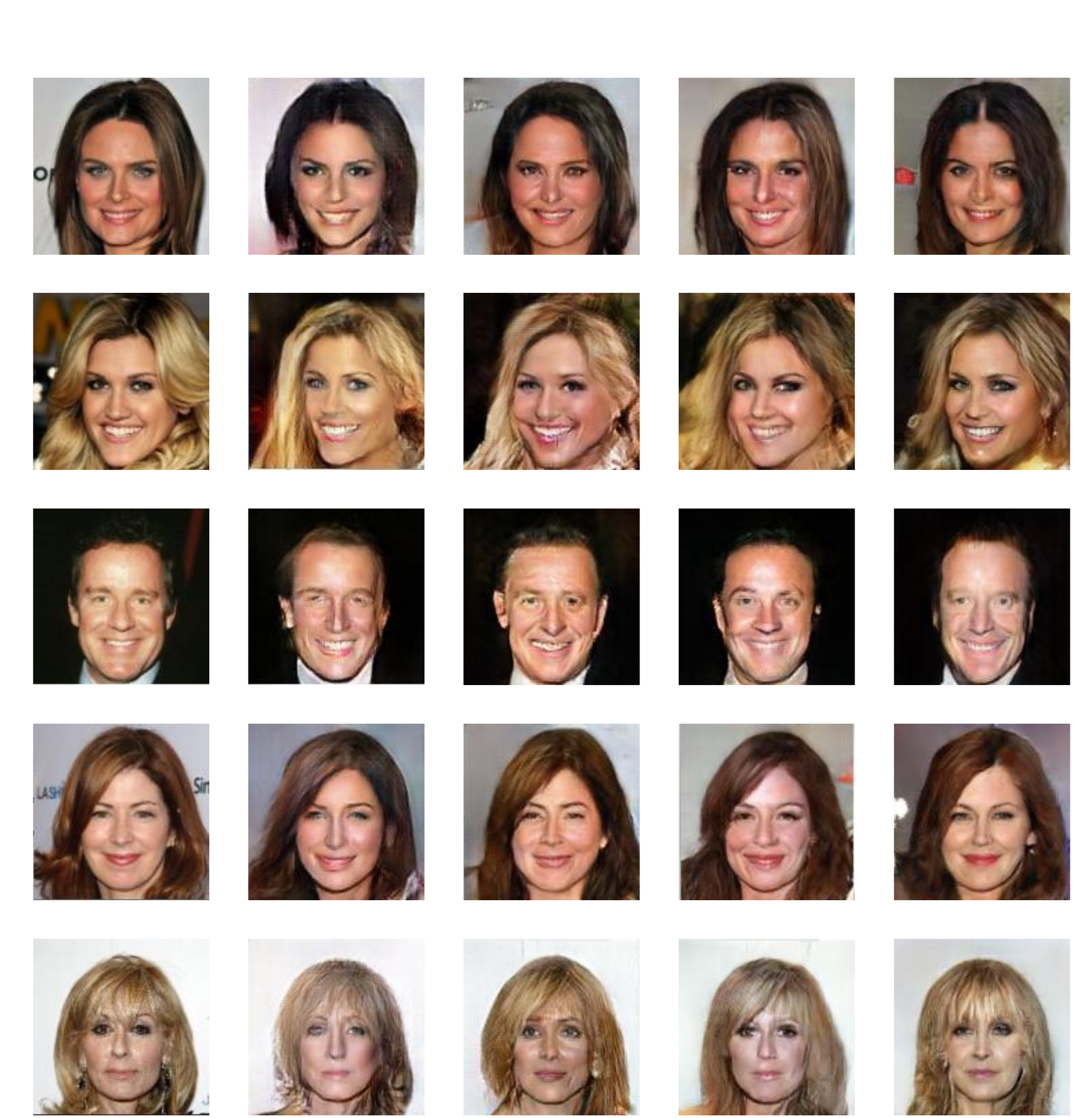}
    \put(6,95){{\footnotesize Target}}
    \put(21,95){{\footnotesize per-pixel\_loss}}
    
    \put(45,95){{\footnotesize $\vz$\_loss}}
    
    \put(58,95){{\footnotesize per-pixel\_loss}}
    \put(68,97.5){{\footnotesize +}}
    \put(65,100){{\footnotesize $\vz$\_loss}}
    
    \put(80,98){{\footnotesize integrated }}
     \put(84,95){{\footnotesize loss}}
    
    \put(25,74.5){{\scriptsize 0.8669}}
    \put(45,74.5){{\scriptsize 0.8220}}
    \put(64,74.5){{\scriptsize 0.8857}}
    \put(83,74.5){{\scriptsize \textbf{0.5314}}}
    
    \put(25,55){{\scriptsize 0.7855}}
    \put(45,55){{\scriptsize 0.9658}}
    \put(64,55){{\scriptsize 0.7200}}
    \put(83,55){{\scriptsize \textbf{0.6089}}}
    
    \put(25,36){{\scriptsize 0.8521}}
    \put(45,36){{\scriptsize 0.7803}}
    \put(64,36){{\scriptsize 0.6086}}
    \put(83,36){{\scriptsize \textbf{0.5217}}}
    
    \put(25,17){{\scriptsize 0.9038}}
    \put(45,17){{\scriptsize 0.9311}}
    \put(64,17){{\scriptsize 0.7412}}
    \put(83,17){{\scriptsize \textbf{0.7259}}}
    
    \put(25,-2.5){{\scriptsize 0.7041}}
    \put(45,-2.5){{\scriptsize 0.8588}}
    \put(64,-2.5){{\scriptsize 0.6779}}
    \put(83,-2.5){{\scriptsize \textbf{0.4636}}}

\end{overpic}
\caption{Comparison of rebuilding images through different losses. The first column shows the original nature-image. The 2nd and 3rd columns are images rebuilt with only per-pixel loss or $\vz$\_loss (latent code). Column 4 shows the effect of $\vz$\_loss + pixel\_loss effect. The last column shows the final effect of the integrated loss. The FaceNet scores below each image in 2nd to 5th columns reveals the distance from the target image. Images rebuild from integrated loss(the last column) get the smallest score(expressed in black font). 
}
\label{fig:lossEvaluation}
\end{figure}

The comparisons among the results of the 4 experiments are shown in Fig. \ref{fig:lossEvaluation}. The First column displays the original images to be rebuilt. The remains are corresponding to the 4 groups of experiments mentioned before. 

From column 2 to column 5, we have three visual observations:
\begin{itemize}
    \item latent code loss ($\vz$\_loss) prefers to preserve image quality (images reconstructed with  per-pixel loss in column 2 are blurrier than images reconstructed with only the latent code loss in the 3rd column) because the calculation of per-pixel loss over the whole pixel set flattens the image; 
    \item images reconstructed with per-pixel loss, latent code loss, latent code loss + per-pixel loss all failed on preserving the face identity;
    \item the integrated loss benefits the effects of its three components that reconstruct the original faces with high quality and identity-preserving.
\end{itemize}

FaceNet also calculates an identity-preserving score for each rebuild face as shown in Fig.  \ref{fig:lossEvaluation} (from column 2 to column 5). A smaller score indicates a closer relationship between two images. The scores from column 2 to column 5 demonstrate that the faces reconstructed by the integrated loss preserve better facial identity than the faces rebuild with other losses (column 2 to 4) in most of the cases. In other words, the integrated loss not only has an advantage on produce high-quality image but also can make a good facial identity-preserving.
These experiments prove that the iterative GAN provided with integrated loss has a strong ability to deal with the tasks mentioned above and archives similar or better results than previous work.


\subsubsection{Face Transformation with Controllable Attributes}
Based on our framework, we feed the discriminator $\mD$ a real image without its attribute labels $\vc$, then we get a noise representation $\tilde{\vz}$ and label vector $\tilde{\vc}$ from $\mC$ as output. We can reconstruct the original image with $\tilde{\vz}$ and $\tilde{\vc}$ as we did in Section \ref{sec:reconstrction}. 
Alternatively, we can transform the original image into another one by customizing its attributes in $\tilde{\vc}$. By modifying part or even all labels of $\tilde{\vc}$, the corresponding attributes of the reconstructed image will be transformed.

\begin{figure*} 
\centering
\begin{overpic}[scale=0.6]{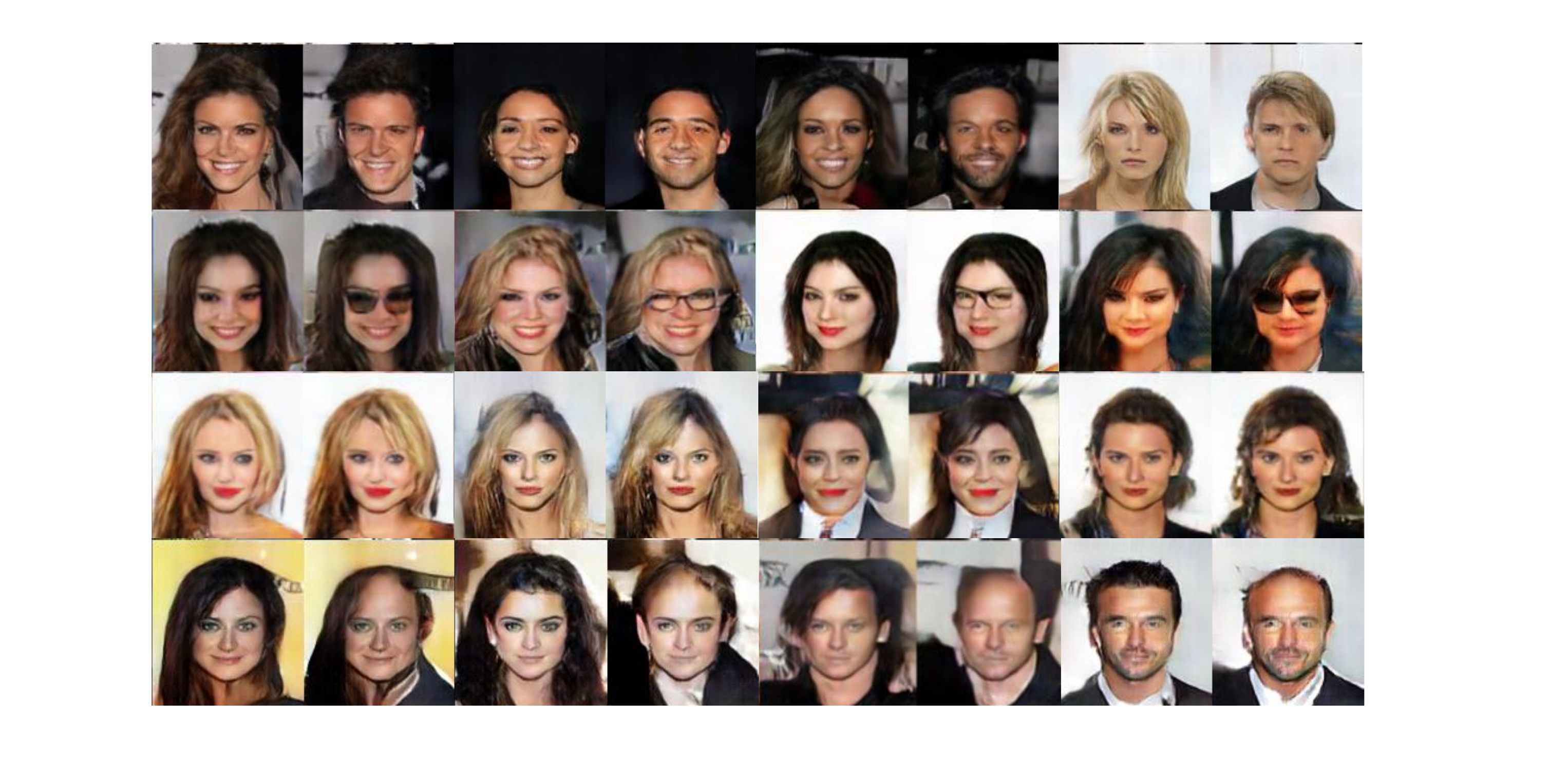}
    \put(3,39){{\footnotesize Male}}
    \put(2,29){{\footnotesize Glasses}}
    \put(2,19){{\footnotesize Bangs}}
    \put(3,8.5){{\footnotesize Bald}}
\end{overpic}
\caption{Four examples (male, glasses, bangs, and bald as shown in the four rows) of face transformation with a single attribute changing. For each example (row), we display five illustrations. For instance, the first row shows the results of controlling the label of the male. The odd columns of row 1 are the given faces, the corresponding even columns display the faces with male reversed. From the 1st and 5th instances (columns 1,2 and 9,10), we clearly see that the mustache disappeared from male to female.
}
\label{fig:singleLabel}
\end{figure*}

In this section, we study the performance of image transformation of our model. We begin the experiments with controlling a single attribute. That is, to modify one of the attributes of the images on the test set. Fig. \ref{fig:singleLabel} shows the part of the results of the transformation on the test set. The four rows of Fig. \ref{fig:singleLabel} illustrate four different attributes \emph{\textbf{Male}}, \emph{\textbf{Eye\_Glasses}}, \emph{\textbf{Bangs}}, \emph{\textbf{Bald}} have been changed. The odd columns display the original images, and even columns illustrate the transformed images. We observe that the transformed images preserve high fidelity and their old attributes.

\begin{figure*} 
\centering
\begin{overpic}[scale=0.55]{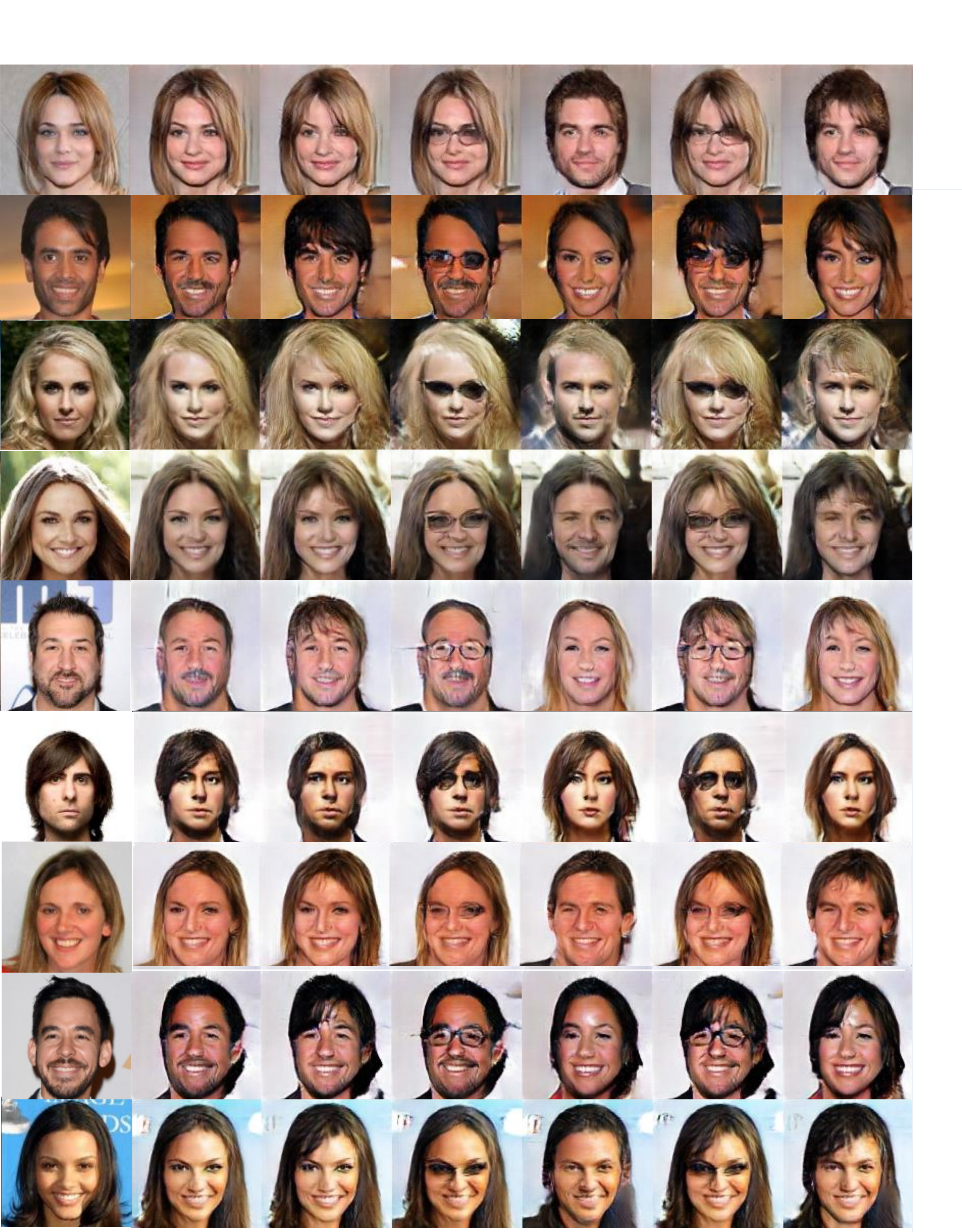}
    \put(2.5,95.3){{\tiny Target}}
    \put(12.5,95.3){{\tiny Rebuild}}
    \put(21.5,95.3){{\tiny Reversal:Bangs}}
    \put(31.5,95.3){{\tiny Reversal:Glasses}}
    \put(42.5,95.3){{\tiny Reversal:Male}}
    \put(53.7,95.3){{\tiny Reversal:Bangs}}
    \put(53.7,96.6){{\tiny Reversal:Glasses}}
    \put(64.6,95.3){{\tiny Reversal:Male}}
    \put(64.6,96.6){{\tiny Reversal:Bangs}}
\end{overpic}
\caption{Demonstrations of rebuilding the target images and reversing (modifying) some attributes of the original face simultaneously. The first column shows the target face. The rebuilt faces are shown in the 2nd column with all attributes unchanged. Then we reverse the 3 single labels successively from 3rd column to the 7th column. For example, the target face has the attribute 'Bangs' (column 3), we reverse the corresponding label 'Bangs' to eliminate this attribute and keep others fixed. The last 2 columns show the combination of attributes modification. }
\label{fig:Multi-label_transfromation}
\end{figure*}

Finally, we extend it to the attribute manipulation from the single case to the multi-label scenario.
Fig. \ref{fig:Multi-label_transfromation} exhibits the results manipulating multiple attributes. 
The first column is the target faces. Faces in column 2 are the corresponding ones that reconstructed by iterative GAN (no attributes have been modified). The remaining 5 columns display the face transformation (column 3,4,5: single attribute; column 6,7: multiple attributes). For these transformed faces, we observe that both the image quality and face identity preserving are well satisfied. As we see, for the multi-label case, we only modified 3 attributes in the test.  Actually, we have tried to manipulate more attributes (4-6) one time while the image quality drastically decreased. There is a lot of things to improve and we left it as the future work.


\subsubsection{Compare with Existing Method of Image Transformation}
Image transformation puts the emphases on finding a way to map the original image to an output image which subjects to another different domain. Editing the image attributes of a person is a special topic in this area. 
One of the popular methods attracted lots of attention recently is the CycleGAN~\cite{Zhu2017Unpaired}.
The key point of CycleGAN is that it builds upon the power of the PIX2PIX~\cite{denoord2016conditional} architecture, with discrete, unpaired collections of training images. 

In this experiment, we compare CycleGAN with iterative GAN on face transformation. We randomly select three facial attributes the \emph{\textbf{Bangs}} ,\emph{\textbf{Glasses}}, and \emph{\textbf{Bald}} for testing. 

For CycleGAN, we split the training dataset into 2 groups for each of the three attributes. For example, to train CycleGAN with the \emph{\textbf{Bangs}}, we divide the images into 2 sets,  faces with bangs belong to domain 1 and the ones without bangs belong to domain 2.
According to the results shown in Figure  \ref{fig:Compare-with-other-gans} $(a)$, we found that CycleGAN is insensitive to the geometry transformation though it did a good job in catching some different features between two domains like color. As we know, CycleGAN is good at transforming the style of an image, for example, translate a horse image to zebra one~\cite{Zhu2017Unpaired}. For the test of human faces, however, it fails to recognize and manipulate the three facial attributes \emph{\textbf{Bangs}} ,\emph{\textbf{Glasses}}, and \emph{\textbf{Bald}} as shown in column 2 of Fig. \ref{fig:Compare-with-other-gans} $(a)$.
By contrast, the iterative GAN achieves better results in transforming the same face with one attribute changed and others preserved.

\subsubsection{Face Generation with Controllable Attributes}
Different from above, we can also generate a new face with a random $\vz$ sampling from a given distribution and an artificial attribute description $\vc$ (labels). The generator $\mG$ accepts $\vz$ and $\vc$ as the inputs and fabricates a fictitious facial image $\vx_{\text{fake}}$ as the output. Of course, we can customize an image by modifying the corresponding attribute descriptions in $\vc$.  For example, the police would like to get a suspect's portrait by the witness's description {\bfseries ``He is around 40 years old bald man with arched eyebrows and big nose"}.  
\begin{figure*} 
\centering
\begin{overpic}[scale=0.5]{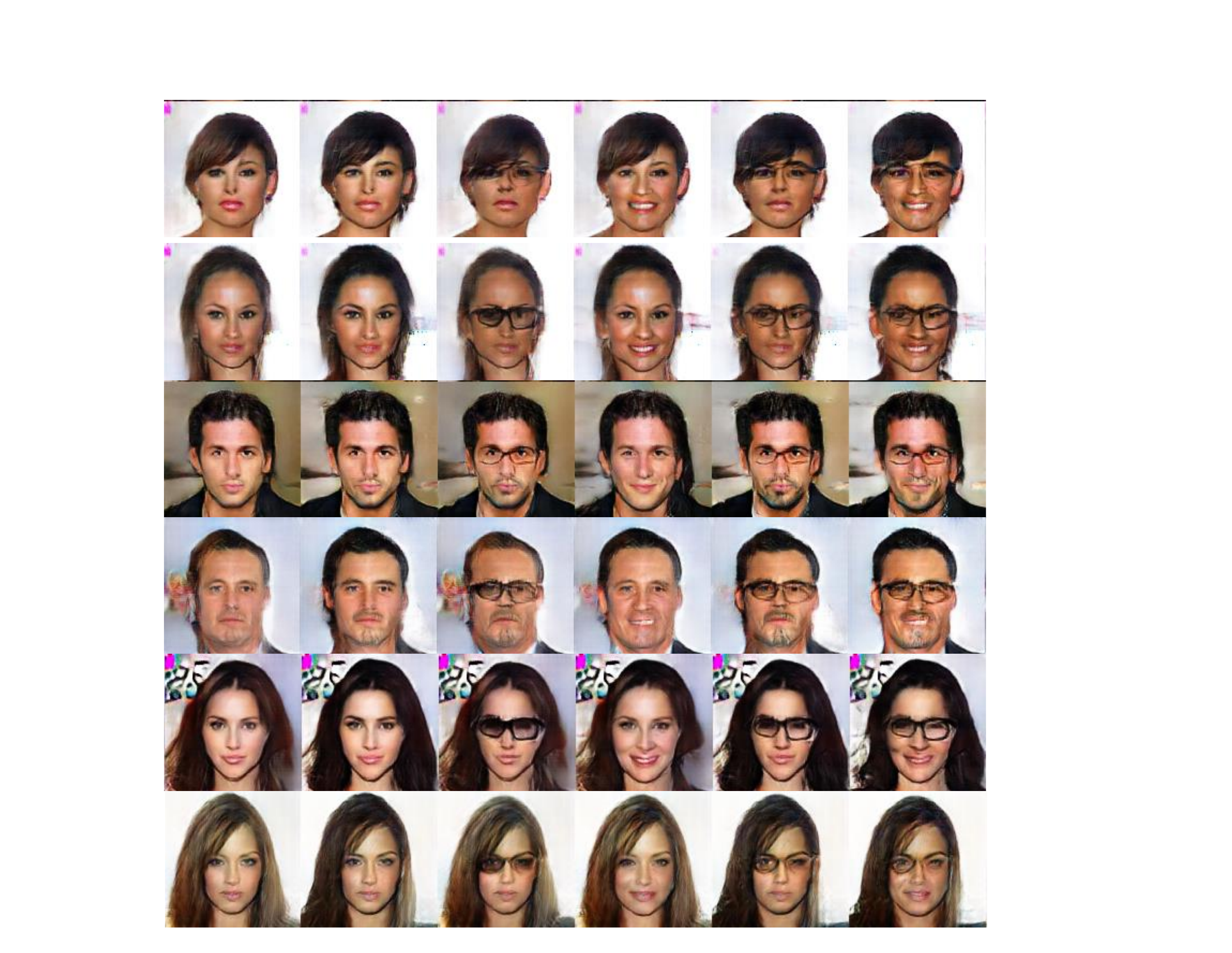}
    \put(-3,35.4){{\scriptsize Random build from}}
    \put(0,33.5){{\scriptsize $\vz \sim U(-1,1)$}}
    
    
    \put(24,70){{\tiny +Bush\_Eyebrows}}
    
    \put(38,70){{\tiny +Glasses}}
    
    \put(48,70){{\tiny +Smiling}}
    
    \put(58,70){{\tiny +Bangs}}
    \put(58,71.8){{\tiny +Bush\_Eyebrows}}
    
    \put(70,70){{\tiny +Bangs}}
    \put(70,71.8){{\tiny +Bush\_Eyebrows}}
    \put(70,73.6){{\tiny +Glasses}}

    
    
\end{overpic}
\caption{Demonstration of building faces.The 1st column is faces build from random noise $\vz$ which is sampled from the Uniform distribution. The 2nd to 4th columns show the standard faces created with the noise vector $\vz$ plus single label such as Bush\_Eyebrows, Glasses, Smiling. The left 2 columns are the examples of manipulating multiple attributes.}
\label{fig:Multi-label_generation}
\end{figure*}
Fig. \ref{fig:Multi-label_generation} illustrates the results of generating fictitious facial images with random noise and descriptions. We sample $\vz$ from the Uniform distribution. The first column displays the images generated with $\vz$ and initial descriptions. The remaining columns demonstrate the facial images generated with modified attributes (single or multiple modifications).

\subsubsection{Compare with Existing Method of Face Generation}
\begin{figure*}
\centering
\begin{overpic}[scale=0.48]{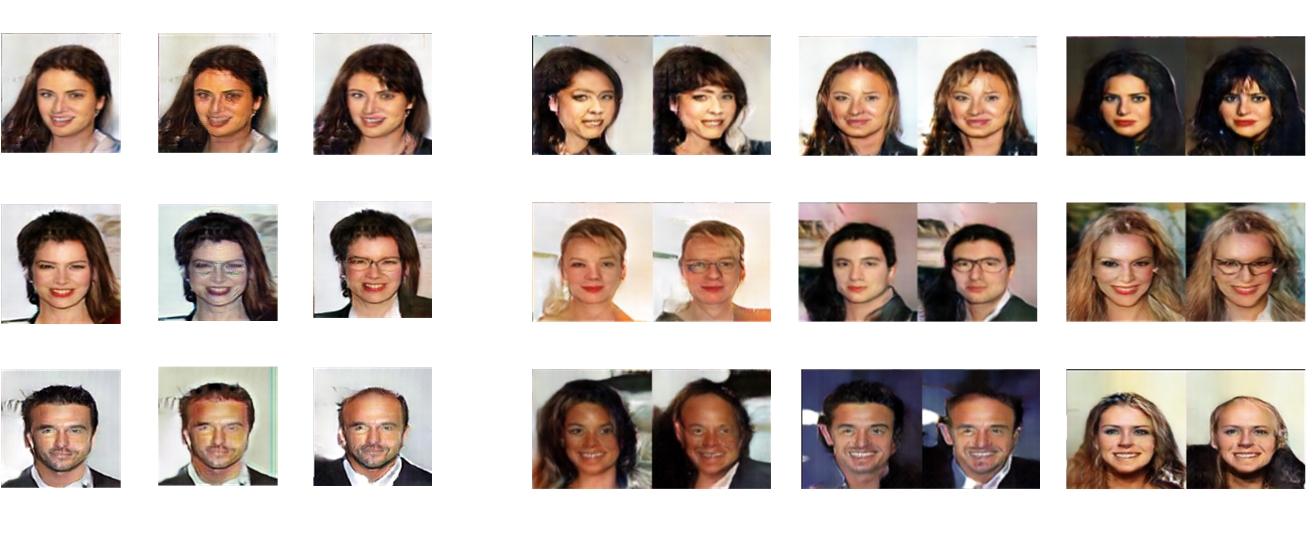}
    \put(2,40){{\footnotesize Target}}
    \put(12,40){{\footnotesize CycleGAN}}
    \put(22.5,40){{\footnotesize iterative GAN}}
   
    \put(46,40){{\footnotesize CGAN}}
    \put(66,40){{\footnotesize AC-GAN}}
    \put(84,40){{\footnotesize iterative GAN}}
    
    \put(-6,34){{\footnotesize Bangs}}
    \put(-6.5,20.5){{\footnotesize Glasses}}
    \put(-5,8){{\footnotesize Bald}}
    
    \put(15,27){{\scriptsize 0.6749 }}
    \put(26,27){{\textbf{0.5520} }}
    
    \put(48,27){{\scriptsize 0.7432 }}
    \put(68,27){{\scriptsize 1.0794 }}
    \put(88,27){{\textbf{0.6066}  }}
    
    \put(15,14.5){{\textbf{0.4480} }}
    \put(26,14.5){{\scriptsize 0.5871 }}
    
    \put(48,14.5){{\scriptsize 1.2572 }}
    \put(68,14.5){{\scriptsize 0.5520 }}
    \put(88,14.5){{\textbf{0.5128} }}
    
    \put(15,1.7){{\scriptsize 0.6112 }}
    \put(26,1.7){{\textbf{0.6069} }}
   
    \put(48,1.7){{\scriptsize 0.7831 }}
    \put(68,1.7){{\scriptsize 0.6928 }}
    \put(88,1.7){{\textbf{0.5434} }}
    
    
    \put(21,-1){{\footnotesize $(a)$}}
    \put(70,-1){{\footnotesize $(b)$}}
    
\end{overpic}
 \caption{Comparing with iterative GAN and other GANs on face generation and transformation. Part $(a)$ shows the results of transforming given images to another with facial attributes changed. When Cycle GAN is performed,the quality of output is so poor since Cycle GAN seems to insensitive to the subtle changes of facial attributes and even failed in some labels(see the image in row 3 column 2). Part $b$ compared the capability of generating an image with iterative GAN and CGAN, AC-GAN.
 The result shows that our model can produce the desired image with comparable or even better quality than AC-GAN, and far better than native CGAN. More importantly, according to the FaceNet scores(below each output image), it seems clear that the proposed iterative architecture has an advantage of preserving facial identity.
 }
 \label{fig:Compare-with-other-gans}
\end{figure*}
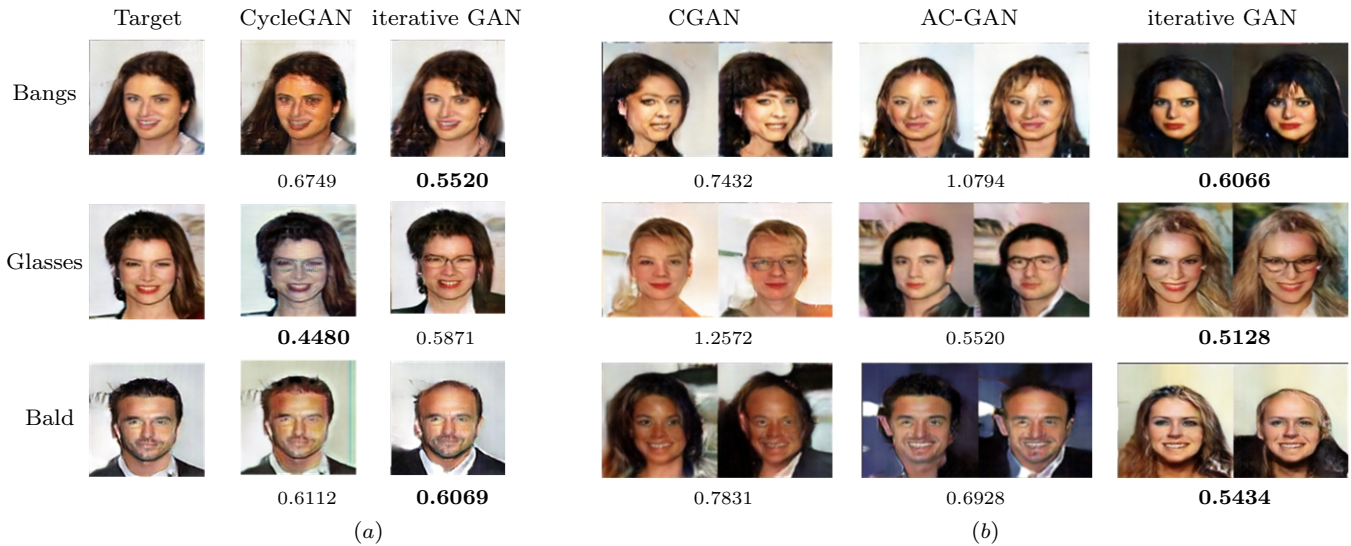

To examine the ability to generate realistic facial images of the proposed model, we compare the results of face generation of the proposed model with two baselines, the CGAN \cite{Mirza2014Conditional} and ACGAN \cite{Odena2017Conditional} respectively.
These three models can all generate images with a conditioned attributes description. For each of them, we begin the experiment by generating random facial images as illustrated in the 1st, 3rd, and 5th column of Fig. \ref{fig:Compare-with-other-gans}, part $(b)$, respectively. The column 2, 4, and 6 display the generated images with the 3 attributes (\emph{\textbf{Bangs}},  \emph{\textbf{Glasses}}, \emph{\textbf{Bald}} modified for CGAN, ACGAN, and our model. It is clear to see that the face quality of our model is better than CGAN and ACGAN. And most importantly, in contrast with the failure of preserving face identity (see Fig. \ref{fig:Compare-with-other-gans} $(b)$, the intersections between column 3, 4 and row 1 of ACGAN, column 1, 2 and row 2, 3 of CGAN), our model can always perform the best in face identity preserving.

It is clear to see that the face quality of ACGAN and our model is much better than CGAN. And most importantly, in contrast with the failure of preserving face identity (see the intersections between column 3, 4 and row 1, 2 of ACGAN), our model can always make a good face identity-preserving.

In summary, extensive experimental results indicate that our method is capable of:
1)recognizing facial attribute; 2)generating high-quality face images with multiple controllable attributes; 3)transforming an input face into an output one with desired attributes changing; 4)preserving the facial identity during face generation and transformation.

\section{Conclusion}
\label{sec:conclusion}
We propose an iterative GAN to perform face generation and transformation jointly by utilizing the strong dependency between the face generation and transformation. To preserve facial identity, an integrated loss including both the per-pixel loss and the perceptual loss is introduced in addition to the traditional adversarial loss. 
Experiments on a real-world face dataset demonstrate the advantages of the proposed model on both generating high-quality images and transforming image with controllable attributes. 

\begin{acknowledgements}
This work was partially supported by the Natural Science Foundation of China (61572111, G05QNQR004), the National High Technology Research and Development Program of China (863 Program) (No. 2015AA015408), a 985 Project of UESTC (No.A1098531023601041) and a Fundamental Research Fund for the Central Universities of China (No. A03017023701012). 
\end{acknowledgements}

\bibliographystyle{spmpsci}      
\bibliography{reference.bib}   

\begin{thebibliography}{10}
\providecommand{\url}[1]{{#1}}
\providecommand{\urlprefix}{URL }
\expandafter\ifx\csname urlstyle\endcsname\relax
  \providecommand{\doi}[1]{DOI~\discretionary{}{}{}#1}\else
  \providecommand{\doi}{DOI~\discretionary{}{}{}\begingroup
  \urlstyle{rm}\Url}\fi

\bibitem{berg2013poof:}
Berg, T., Belhumeur, P.N.: Poof: Part-based one-vs.-one features for
  fine-grained categorization, face verification, and attribute estimation pp.
  955--962 (2013)

\bibitem{bettadapura2012face}
Bettadapura, V.: Face expression recognition and analysis: The state of the
  art.
\newblock Tech Report, arXiv:1203.6722  (2012)

\bibitem{bourdev2011describing}
Bourdev, L.D., Maji, S., Malik, J.: Describing people: A poselet-based approach
  to attribute classification pp. 1543--1550 (2011)

\bibitem{branson2010visual}
Branson, S., Wah, C., Schroff, F., Babenko, B., Welinder, P., Perona, P.,
  Belongie, S.J.: Visual recognition with humans in the loop pp. 438--451
  (2010)

\bibitem{cherniavsky2010semi-supervised}
Cherniavsky, N., Laptev, I., Sivic, J., Zisserman, A.: Semi-supervised learning
  of facial attributes in video pp. 43--56 (2010)

\bibitem{cortes1995support-vector}
Cortes, C., Vapnik, V.: Support-vector networks.
\newblock Machine Learning \textbf{20}(3), 273--297 (1995)

\bibitem{dalal2005histograms}
Dalal, N., Triggs, B.: Histograms of oriented gradients for human detection
  \textbf{1}, 886--893 (2005)

\bibitem{denoord2016conditional}
Den~Oord, A.V., Kalchbrenner, N., Vinyals, O., Espeholt, L., Graves, A.,
  Kavukcuoglu, K.: Conditional image generation with pixelcnn decoders.
\newblock neural information processing systems pp. 4790--4798 (2016)

\bibitem{denton2015deep}
Denton, E.L., Chintala, S., Szlam, A., Fergus, R.D.: Deep generative image
  models using a laplacian pyramid of adversarial networks.
\newblock neural information processing systems pp. 1486--1494 (2015)

\bibitem{dong2014learning}
Dong, C., Loy, C.C., He, K., Tang, X.: Learning a deep convolutional network
  for image super-resolution pp. 184--199 (2014)

\bibitem{dosovitskiy2016generating}
Dosovitskiy, A., Brox, T.: Generating images with perceptual similarity metrics
  based on deep networks.
\newblock neural information processing systems pp. 658--666 (2016)

\bibitem{Farhadi2009Describing}
Farhadi, A., Endres, I., Hoiem, D., Forsyth, D.: Describing objects by their
  attributes.
\newblock In: Computer Vision and Pattern Recognition, 2009. CVPR 2009. IEEE
  Conference on, pp. 1778--1785 (2009)

\bibitem{gatys2015a}
Gatys, L.A., Ecker, A.S., Bethge, M.: A neural algorithm of artistic style.
\newblock Nature Communications  (2015)

\bibitem{gauthier14}
Gauthier, J.: Conditional generative adversarial nets for convolutional face
  generation.
\newblock Class Project for Stanford CS231N: Convolutional Neural Networks for
  Visual Recognition  (Winter semester 2014)

\bibitem{goodfellow2014generative}
Goodfellow, I.J., Pougetabadie, J., Mirza, M., Xu, B., Wardefarley, D., Ozair,
  S., Courville, A.C., Bengio, Y.: Generative adversarial nets pp. 2672--2680
  (2014)

\bibitem{gregor2015draw}
Gregor, K., Danihelka, I., Graves, A., Rezende, D.J., Wierstra, D.: Draw: A
  recurrent neural network for image generation.
\newblock international conference on machine learning pp. 1462--1471 (2015)

\bibitem{Isola2016Image}
Isola, P., Zhu, J.Y., Zhou, T., Efros, A.A.: Image-to-image translation with
  conditional adversarial networks  (2016)

\bibitem{johnson2016perceptual}
Johnson, J., Alahi, A., Feifei, L.: Perceptual losses for real-time style
  transfer and super-resolution.
\newblock european conference on computer vision pp. 694--711 (2016)

\bibitem{adam2015ICLR}
Kingma, D., Ba, J.: Adam: A method for stochastic optimization.
\newblock In: International Conference on Learning Representation (2015)

\bibitem{Kingma2014Semi}
Kingma, D.P., Rezende, D.J., Mohamed, S., Welling, M.: Semi-supervised learning
  with deep generative models.
\newblock Advances in Neural Information Processing Systems \textbf{4},
  3581--3589 (2014)

\bibitem{kingma2014auto-encoding}
Kingma, D.P., Welling, M.: Auto-encoding variational bayes.
\newblock international conference on learning representations  (2014)

\bibitem{kumar2009attribute}
Kumar, N., Berg, A.C., Belhumeur, P.N., Nayar, S.K.: Attribute and simile
  classifiers for face verification pp. 365--372 (2009)

\bibitem{Larsen2015Autoencoding}
Larsen, A.B.L., Sonderby, S.K., Larochelle, H., Winther, O.: Autoencoding
  beyond pixels using a learned similarity metric.
\newblock international conference on machine learning pp. 1558--1566 (2016)

\bibitem{ledig2016photo-realistic}
Ledig, C., Theis, L., Huszar, F., Caballero, J., Cunningham, A., Acosta, A.,
  Aitken, A.P., Tejani, A., Totz, J., Wang, Z., et~al.: Photo-realistic single
  image super-resolution using a generative adversarial network.
\newblock computer vision and pattern recognition pp. 4681--4690 (2016)

\bibitem{li2017generate}
Li, Z., Luo, Y.: Generate identity-preserving faces by generative adversarial
  networks.
\newblock arXiv preprint arXiv:1706.03227  (2017)

\bibitem{liu2015faceattributes}
Liu, Z., Luo, P., Wang, X., Tang, X.: Deep learning face attributes in the
  wild.
\newblock In: Proceedings of International Conference on Computer Vision
  (ICCV), pp. 3730--3738 (2015)

\bibitem{mcquistonsurrett2006use}
Mcquistonsurrett, D., Topp, L.D., Malpass, R.S.: Use of facial composite
  systems in us law enforcement agencies.
\newblock Psychology Crime \& Law \textbf{12}(5), 505--517 (2006)

\bibitem{Mirza2014Conditional}
Mirza, M., Osindero, S.: Conditional generative adversarial nets.
\newblock Computer Science pp. 2672--2680 (2014)

\bibitem{nilsback2008automated}
Nilsback, M., Zisserman, A.: Automated flower classification over a large
  number of classes pp. 722--729 (2008)

\bibitem{Odena2017Conditional}
Odena, A., Olah, C., Shlens, J.: Conditional image synthesis with auxiliary
  classifier gans  (2017)

\bibitem{parikh2011relative}
Parikh, D., Grauman, K.: Relative attributes  (2011)

\bibitem{radford2016unsupervised}
Radford, A., Metz, L., Chintala, S.: Unsupervised representation learning with
  deep convolutional generative adversarial networks.
\newblock International Conference on Learning Representations  (2016)

\bibitem{schroff2015facenet:}
Schroff, F., Kalenichenko, D., Philbin, J.: Facenet: A unified embedding for
  face recognition and clustering.
\newblock computer vision and pattern recognition pp. 815--823 (2015)

\bibitem{ShuYHSSS17}
Shu, Z., Yumer, E., Hadap, S., Sunkavalli, K., Shechtman, E., Samaras, D.:
  Neural face editing with intrinsic image disentangling.
\newblock CoRR \textbf{abs/1704.04131} (2017).
\newblock \urlprefix\url{http://arxiv.org/abs/1704.04131}

\bibitem{simonyan2015very}
Simonyan, K., Zisserman, A.: Very deep convolutional networks for large-scale
  image recognition.
\newblock international conference on learning representations  (2015)

\bibitem{sun2014deep}
Sun, Y., Chen, Y., Wang, X., Tang, X.: Deep learning face representation by
  joint identification-verification.
\newblock neural information processing systems pp. 1988--1996 (2014)

\bibitem{szegedy2015going}
Szegedy, C., Liu, W., Jia, Y., Sermanet, P., Reed, S.E., Anguelov, D., Erhan,
  D., Vanhoucke, V., Rabinovich, A.: Going deeper with convolutions.
\newblock computer vision and pattern recognition pp. 1--9 (2015)

\bibitem{tatarchenko2016multi-view}
Tatarchenko, M., Dosovitskiy, A., Brox, T.: Multi-view 3d models from single
  images with a convolutional network.
\newblock european conference on computer vision pp. 322--337 (2016)

\bibitem{Wang2017Tag}
Wang, C., Wang, C., Xu, C., Tao, D.: Tag disentangled generative adversarial
  network for object image re-rendering.
\newblock In: Twenty-Sixth International Joint Conference on Artificial
  Intelligence, pp. 2901--2907 (2017)

\bibitem{Wang2017Perceptual}
Wang, C., Xu, C., Wang, C., Tao, D.: Perceptual adversarial networks for
  image-to-image transformation  (2017)

\bibitem{wang2016generative}
Wang, X., Gupta, A.: Generative image modeling using style and structure
  adversarial networks.
\newblock european conference on computer vision pp. 318--335 (2016)

\bibitem{xie2012image}
Xie, J., Xu, L., Chen, E.: Image denoising and inpainting with deep neural
  networks pp. 341--349 (2012)

\bibitem{yan2015attribute2image}
Yan, X., Yang, J., Sohn, K., Lee, H.: Attribute2image: Conditional image
  generation from visual attributes.
\newblock european conference on computer vision pp. 776--791 (2015)

\bibitem{yoo2016pixel-level}
Yoo, D., Kim, N., Park, S., Paek, A.S., Kweon, I.S.: Pixel-level domain
  transfer.
\newblock european conference on computer vision pp. 517--532 (2016)

\bibitem{zhang2014panda:}
Zhang, N., Paluri, M., Ranzato, M., Darrell, T., Bourdev, L.D.: Panda: Pose
  aligned networks for deep attribute modeling.
\newblock computer vision and pattern recognition pp. 1637--1644 (2014)

\bibitem{zhang2016colorful}
Zhang, R., Isola, P., Efros, A.A.: Colorful image colorization.
\newblock european conference on computer vision pp. 649--666 (2016)

\bibitem{zhang2017age}
Zhang, Z., Song, Y., Qi, H.: Age progression/regression by conditional
  adversarial autoencoder.
\newblock arXiv preprint arXiv:1702.08423  (2017)

\bibitem{zhu2016generative}
Zhu, J., Krahenbuhl, P., Shechtman, E., Efros, A.A.: Generative visual
  manipulation on the natural image manifold.
\newblock european conference on computer vision pp. 597--613 (2016)

\bibitem{Zhu2017Unpaired}
Zhu, J.Y., Park, T., Isola, P., Efros, A.A.: Unpaired image-to-image
  translation using cycle-consistent adversarial networks  (2017)

\end{thebibliography}


\end{document}